\documentclass[11pt, journal, draftclsnofoot, onecolumn]{IEEEtran}

\usepackage[hyphens]{url}  
\usepackage{graphicx} 

\usepackage{cite}
\usepackage{hyperref}
\usepackage{caption} 
\usepackage{algorithm}
\usepackage{algorithmic}
\usepackage{subcaption}

\usepackage{amsmath,amssymb}
\usepackage{color}
\usepackage{makecell}
\usepackage{amsthm}

\usepackage{mathtools,enumitem,dsfont}

\DeclarePairedDelimiter\floor{\lfloor}{\rfloor}

\newtheorem{Theorem}{Theorem}
\newtheorem{Proposition}{Proposition}

\title{CorBin-FL: A Differentially Private Federated Learning Mechanism using Common Randomness}

\author{%
  \IEEEauthorblockN{Hojat Allah Salehi\IEEEauthorrefmark{1},
                Md Jueal Mia\IEEEauthorrefmark{1}, S. Sandeep Pradhan\IEEEauthorrefmark{2}, 
                 M. Hadi Amini\IEEEauthorrefmark{1},
                    Farhad Shirani\IEEEauthorrefmark{1}
                   }
\\\IEEEauthorblockA{\IEEEauthorrefmark{1}%
Knight Foundation School of Computing and Information Sciences,
\\ Florida International University, Miami, FL,
                     \{hsalehi,mmia001,moamini,fshirani\}@fiu.edu}
\\  \IEEEauthorblockA{\IEEEauthorrefmark{2}%
               Electrical Engineering and Computer Science Department\\University of Michigan, Ann Arbor, MI, pradhanv@umich.edu}
}

\begin{document}

\maketitle

\begin{abstract}
Federated learning (FL) has emerged as a promising framework for distributed machine learning. 
It enables collaborative learning among multiple clients, utilizing distributed data and computing resources. However, FL faces challenges in balancing privacy guarantees, communication efficiency, and overall model accuracy. In this work, we introduce CorBin-FL, a privacy mechanism that uses correlated binary stochastic quantization to achieve differential privacy while maintaining overall model accuracy. The approach uses secure multi-party computation techniques to enable clients to perform correlated quantization of their local model updates without compromising individual privacy.
We provide theoretical analysis showing that CorBin-FL achieves parameter-level local differential privacy (PLDP), and that it asymptotically optimizes the privacy-utility trade-off between the mean square error utility measure and the PLDP privacy measure. We further propose AugCorBin-FL, an extension that, in addition to PLDP, achieves user-level and sample-level central differential privacy guarantees. For both mechanisms, we derive bounds on privacy parameters and mean squared error performance measures.
Extensive experiments on MNIST and CIFAR10 datasets demonstrate that our mechanisms outperform existing differentially private FL mechanisms, including Gaussian and Laplacian mechanisms, in terms of model accuracy under equal PLDP privacy budgets\footnote{For reproducibility, our code is included in the supplementary material and will be made publicly available after publication.}.
\end{abstract}

\section{Introduction}
Distributed stochastic gradient descent in general, and 
federated learning (FL) in particular,  are foundational concepts in modern machine learning with wide-ranging applications
\cite{mcdonald2010distributed,dean2012large,konevcny2016federated}. 
The typical FL setup consists of a collection of distributed clients collaborating with a central server over multiple rounds of communication to train a global model.
Each client is equipped with a local dataset. At each communication round, it
receives a copy of the global model and updates it using local data. The updates, usually gradients, are sent to a server, which aggregates them and updates the global model. This enables privacy-preserving collaborative learning among multiple clients, utilizing distributed data and computing resources, without requiring the clients to disclose sensitive data. The latter property is of particular interest in sensitive applications such as training over health datasets \cite{roy2019braintorrent,xu2021fedmood}.

Federated learning inherently provides a degree of privacy by not requiring clients to share raw data. This is further enhanced when combined with secure aggregation techniques  \cite{bonawitz2017practical} which enable the clients to send the aggregate model to the server without revealing
information about local update values.
However, recent works have shown that both the aggregated model updates at each round and the final model weights may leak information about the training data, and even potentially enable the reconstruction of sensitive data \cite{papernot2016semi,bhowmick2018protection,carlini2018secret,zhu2019deep}.

To provide quantifiable guarantees against data leakage, the concept of differential privacy (DP), introduced by \cite{dwork2006our}, has emerged as a crucial framework. At a high level, DP requires that the output of an (aggregation) mechanism remains essentially unchanged when any single individual's data is added to or removed from the input dataset. Several different notions of DP have been considered in FL. User-level central DP (UCDP) requires that the aggregate model at each round does not reveal the participation of any individual client in that round of training \cite{dwork2014algorithmic,abadi2016deep,mcmahan2017learning}. Sample-level central DP (SCDP) is more stringent and 
requires that the aggregate model does not reveal information about usage of any single training sample \cite{zhao2024learning,ghazi2024user}. Parameter-level local DP (PLDP) requires that an eavesdropper gaining access to the local update being transmitted by one of the clients in uplink communication does not gain information about each individual model parameter \cite{sun2020ldp}. Prior works have considered injecting noise into the updates to achieve DP. These include the Gaussian mechanism, Laplacian mechanism, and the Discrete Gaussian mechanism, among others \cite{bassily2014private,abadi2016deep,canonne2020discrete}.  

The communication cost of sending the gradients to the server often emerges as a performance bottleneck in FL. This is especially the case in scenarios where the clients are mobile devices with limited uplink bandwidth \cite{agarwal2018cpsgd}. Recent works have focused on techniques such as gradient quantization to reduce the communication overhead \cite{seide20141,suresh2017distributed,chen2024privacy}. In addition to gains in  communication efficiency, quantization may yield improved privacy guarantees.
A stochastic quantization method, called LDP-FL, was introduced in \cite{sun2020ldp}, which, in addition to reducing the communication overhead via one-bit quantization of each parameter update, guarantees PLDP. Alternative mechanisms adding discrete noise to  quantized updates to achieve DP have also been considered, such as the binomial mechanism  in \cite{dwork2006our,agarwal2018cpsgd}.

In this work, we consider correlated stochastic quantization to achieve various notions of DP, such as UCDP, SCDP, and PLDP. A key feature of our framework is that clients use shared common randomness to coordinate while maintaining privacy. To elaborate, at each communication round, the clients use the public channel facilitated by the server to exchange private keys using the Diffie-Hellman (DH) key exchange protocol \cite{diffie1976new}. The application of the DH protocol to construct pairwise secure communication among the clients has been previously explored in the literature, for instance the secure aggregation mechanism \cite{bonawitz2017practical}. After exchanging private keys, clients are partitioned into pairs, and each pair uses their secure channel to share a few bits of common randomness per model parameter. The shared randomness is used to perform correlated stochastic quantization (Algorithm \ref{Alg:algCQ}) of the model parameters, without requiring the clients to share any information about their respective model parameters. The quantizer is designed to achieve PLDP. Furthermore, the quantizer is unbiased and the quantization noises are correlated, such that they cancel each other in the aggregation phase, yielding small mean square error (MSE) between the aggregate of the quantized updates and that of the original local model updates. 
We refer to the resulting privacy mechanism, which uses correlated binary stochastic quantization, as the CorBin-FL mechanism. We also introduce the AugCorBin-FL mechanism, in which a fraction of the clients pair up and perform correlated quantization as in CorBin-FL, and the rest of the clients quantize their model weights individually. We show that the AugCorBin-FL mechanism achieves UCDP, SCDP, and PLDP and quantify the corresponding privacy parameters and achievable MSE. The idea of injecting correlated noise to achieve DP, via  common randomness shared among pairs of clients, has been studied in two concurrent works, where the Gaussian mechanism was considered in FL \cite{allouah2024privacy} and general distributed mean estimation scenarios \cite{vithana2024correlated}. A main advantage of CorBin-FL is that it only requires a few bits of common randomness  to be shared among the clients per model parameter to achieve close to optimal MSE
(Theorem \ref{th:1} and Figure \ref{fig:all_comparisons}(a)). This is in contrast with
the aforementioned mechanisms which consider maximal negative correlation among Gaussian variables, that requires an asymptotically large number of common random bits per model parameter \cite{witsenhausen1975sequences}. Our contributions are summarized as follows:
\begin{itemize}[leftmargin=*]
    \item We introduce a privacy-utility tradeoff, in terms of the MSE utility measure and the PLDP privacy measure, and characterize the associated optimal class of binary stochastic quantizers. (Algorithm \ref{Alg:algQ} and Proposition \ref{prop:1})
    \item We extend the formulation of the privacy-utility tradeoff to correlated stochastic pairs of quantizers, and  provide an algorithm for asymptotically optimal correlated quantization. (Algorithm \ref{Alg:algCQ} and Theorem \ref{th:1})
    \item We introduce the CorBin-FL mechanism (Algorithm \ref{Alg:CorBinFL}), and provide privacy and utility guarantees in terms of PLDP and MSE. (Theorem \ref{prop:util})
    \item We introduce AugCorBin-FL and derive guarantees in terms of PLDP, UCDP, and MSE. (Theorem \ref{th:3}) 
    \item We provide extensive empirical simulations demonstrating performance gains in terms of model accuracy under a fixed privacy budget over the Gaussian mechanism, the Laplacian mechanism, and LDP-FL, and other privacy mechanisms, on the MNIST and CIFAR10 datasets.
\end{itemize}
\textit{Notation:} Sets are denoted by calligraphic letters such as $\mathcal{X}$. 
The set $\{1,2,\cdots, n\}$ is represented by $[n]$. 
Vectors and matrices are denoted by bold-face letters such as $\mathbf{x}$ and $\mathbf{h}$.
The symbol $\prec$ denotes the lexicographic ordering on binary vectors.
For $p\in \mathbb{N}$, we write $||\cdot||_p$ to denote the $L_p$-norm.  
 Upper-case letters such as $X$ represent random variables, and lower-case letters such as $x$ represent their realizations.  Similarly, random vectors and random matrices are denoted by upper-case letters such as $\mathbf{X}$  and $\mathbf{H}$, respectively.

\section{Preliminaries}
\label{Sec:Pre}
\subsection{Private Federated Learning}
In this work, we propose privacy mechanisms and quantify their associated privacy gains when clients share a source of (limited) common randomness. 
The setting comprises $n$ clients, denoted by $\mathcal{C}_i, i\in [n]$, collaborating with a central server $\mathcal{S}$ to train a global model iteratively over multiple communication rounds. Each client $\mathcal{C}_i$ possesses a local dataset $\mathcal{D}_i$. At each communication round, the server sends the current global model to all clients, i.e., it sends the vector $\mathbf{w}_g= (w_{g,j})_{j\in [m]}$, where $m$ is the number of model parameters.  
The $i$th client updates its local model based on $\mathcal{D}_i$ and computes the updated local model parameters $\mathbf{w}_{i}=(w_{i,j})_{j\in [m]}, i\in [n]$.
For any given $j\in [m]$, let the  bounded interval $[c_j-r_j,c_j+r_j]$ be chosen such that it contains the collection of updated local model parameters, $w_{i,j},i\in [n]$. The variable $c_j$ is called 
the \textit{center} and the variable $r_j$ the \textit{radius}\footnote{
The center $c_j$ and radius  $r_j$ depend on the range of $w_{i,j}, i\in [n]$.
In practice, the clients to do not have access to each other's updated parameters. Consequently, the updated parameters are clipped by each client using a $\hat{c}_j$ and $\hat{r}_j$, which are shared by the server at each round to ensure that $w_{i,j}\in [\hat{c}_j-\hat{r_j},\hat{c}_j+\hat{r}_j]$.} corresponding to the $j$th model parameter. 
A key feature of our framework is that clients share a common vector of binary symmetric random variables $\mathbf{Z}^{d\times m}$, where $d$ is the number of shared bits per model parameter. We will describe the method to share this common vector among the clients in the subsequent sections.
Clients apply a (possibly stochastic) privacy mechanism $\mathcal{M}:\mathbb{R}^m \times\mathbb{R}^{d\times m}\to \mathbb{R}^m$, generating the obfuscated vectors $\overline{\mathbf{W}}_i=\mathcal{M}(\mathbf{w}_i,\mathbf{Z}^{d\times m})$, which are transmitted to the server. The server aggregates the obfuscated vectors, computing the updated parameters $\overline{\mathbf{W}}_g=\frac{1}{n}\sum_{i\in [n]}\overline{\mathbf{W}}_i$.
The mechanism $\mathcal{M}$ should satisfy differential privacy constraints, as detailed in subsequent sections, while maintaining overall utility, in terms of the global model accuracy. In our theoretical derivations, the MSE between the aggregate of the obfuscated local updates and that of the unobfuscated updates is used as a surrogate utility measure. Our empirical evaluations show that the mechanism achieves improved accuracy over widely used methods such as the Gaussian and Laplacian mechanisms. 


\subsection{Local and Central Differential Privacy}
We consider the following notions of differential privacy.
\subsubsection{User-Level Central Differential Privacy:} For $\epsilon_u,\delta>0$, a mechanism is said to achieve $(\epsilon_u,\delta)$-UCDP if:
\begin{align}
\label{eq:UCDP}
P(\mathbf{\overline{W}}_g\in \mathcal{T})\leq e^{\epsilon_u} P(\mathbf{\overline{W}'}_g\in \mathcal{T})+\delta,
\end{align}
where $\overline{\mathbf{W}}_g=\frac{1}{n}\sum_{i=1}^n \overline{\mathbf{W}}_i$,$\overline{\mathbf{W}}'_g=\frac{1}{n}\sum_{i=2}^n \overline{\mathbf{W}}_i$, $\overline{\mathbf{W}}_i= \mathcal{M}(\mathbf{{w}}_i, \mathbf{Z}^{{d\times m}}), \mathbf{{w}}_i\in \mathbb{R}^m$, and $\mathcal{T}\subseteq \mathbb{R}^m$. At a high level, this condition guarantees that an adversary having access to the updated aggregate model weights cannot reliably detect if a specific user has participated in the training round. 
\subsubsection{Sample-Level Central Differential Privacy:} For $\epsilon_s,\delta>0$, a mechanism is said to achieve $(\epsilon_s,\delta)$-SCDP if:
\begin{align}
\label{eq:SCDP}
P(\mathbf{\overline{W}}_g\in \mathcal{T})\leq e^{\epsilon_s} P(\mathbf{\overline{W}'}_g\in \mathcal{T})+\delta,
\end{align}
where $\overline{\mathbf{W}}_g=\frac{1}{n}\sum_{i=1}^n \overline{\mathbf{W}}_i$,$\overline{\mathbf{W}}'_g=\frac{1}{n}\overline{\mathbf{W}}'_1+\frac{1}{n}\sum_{i=2}^n \overline{\mathbf{W}}_i$, 
$\overline{\mathbf{W}}_i= \mathcal{M}(\mathbf{{w}}_i, \mathbf{Z}^{d\times m}), \mathbf{{w}}_i\in \mathbb{R}^m$, 
$\overline{\mathbf{W}}'_1= \mathcal{M}(\mathbf{{w}}'_1, \mathbf{Z}^{d\times m}), \mathbf{{w}}'_1\in \mathbb{R}^m$ such that $\|{\mathbf{w}}_1-{\mathbf{w}}'_1\|_2 \leq \Delta_2$, $\Delta_2>0$ is called the sensitivity parameter, and $\mathcal{T}\subseteq \mathbb{R}^m$. This condition, for an appropriately chosen value of $\Delta_2$, guarantees that an adversary having access to the updated aggregate model weights cannot reliably detect if a specific training sample was used in the training process.

\subsubsection{Parameter-Level Local Differential Privacy:} Following \cite{sun2020ldp,truex2020ldp}, 
for $\epsilon_p>0$, a mechanism $\mathcal{M}$ is said to achieve $\epsilon_p$-PLDP if:
\begin{align}
\label{eq:LDP}
\max_{j\in [m]} \frac{P(\overline{W}_{i,j}\in \mathcal{T})}{P(\overline{W}'_{i,j}\in \mathcal{T})}\leq e^{\epsilon_p},
\end{align}
for all $i\in [n]$, where $\mathcal{T}\subseteq \mathbb{R}^m$, $\mathbf{\overline{W}_i}= \mathcal{M}(\mathbf{w}_i,\mathbf{Z}^{d\times m})$ and $\mathbf{\overline{W}'_i}= \mathcal{M}({\mathbf{w}}'_i,\mathbf{Z}^{d\times m})$, and $\mathbf{w}_i,\mathbf{w}'_i\in \mathbb{R}^{m}$. This guarantees the privacy of each local model parameter if the uplink communication between  client and server is compromised.

\subsection{The LDP-FL Mechanism}
In the subsequent sections, we build upon the LDP-FL mechanism \cite{sun2020ldp} to introduce the CorBin-FL and AugCorBin-FL privacy mechanisms. For completeness, we provide a brief summary of the LDP-FL mechanism in this section. 
For a fixed $\epsilon_p>0$,
the mechanism uses a stochastic quantizer, denoted by $\textsc{LDPQ}(\cdot)$, to obfuscate each of the local model parameters. The $\textsc{LDPQ}(\cdot)$ quantizer is described in Algorithm \ref{Alg:algQ}. To provide an overview, given an input weight $w$, center $c$, and radius $r$, the quantizer produces a binary random variable $U$ with 
\[P(U=1)=1-P(U=-1)= \frac{1}{2}+\frac{w-c}{2r\alpha(\epsilon_p)},\] 
where $\alpha(\epsilon_p)=\frac{1+e^{\epsilon_p}}{1-e^{\epsilon_p}}$.
Each client computes $\overline{W}_{i,j}=LDPQ(\epsilon_p,c_j,r_j,w_{i,j}), j\in [m]$ and sends it to the server.

To gain insights into the operation of LDP-FL, let us consider a scenario with no privacy requirements, i.e., $\epsilon_p \to \infty$. Then, $\alpha(\epsilon_p)\to 1$ and if ${w}_{i,j}>c_j$, then $P(U=1)>P(U=-1)$, i.e., it is more likely that the client sends the update $\overline{W}_{i,j}= c_j+r_j$. Conversely, if ${w}_{i,j}<c_j$, then it is more likely that it sends the update $\overline{W}_{i,j}= c_j-r_j$.  This can be interpreted as the client sending the direction of change, rather than the actual value of the updated model parameter.

As the privacy budget $\epsilon_p$ is decreased, the client update becomes less \textit{truthful}, i.e., it becomes more likely that $U$ points to the direction opposite to the true update direction. On average, the \textit{misdirections} of different clients average out since the mechanism produces an unbiased estimate, and the aggregate update on the server-side remains accurate.  To see this, note that from Line 4 in Algorithm \ref{Alg:algQ}, we have $\mathbb{E}(U)= \frac{{w}_{i,j}}{2r_j\alpha(\epsilon_p)}$, so that $\mathbb{E}(\overline{W}_{i,j})= {w}_{i,j}$.  When the number of participating clients is sufficiently large, the unbiased property ensures that the aggregate of the obfuscated parameters closely approximates that of the original, unobfuscated inputs. The MSE of the parameter estimate is $\mathbb{E}((\overline{W}_{g,j}-w_{g,j})^2) \approx \frac{r_j^2 \alpha^2(\epsilon_p)}{n}$ \cite[Lemma 3]{sun2020ldp}, and vanishes as $n$ becomes asymptotically large. 

\section{Correlated Stochastic Quantizers}
In this section, we introduce a novel class of correlated stochastic quantizers.
We first show that the $\textsc{LDPQ}$ quantizer is the solution to a specific privacy-utility optimization. Then, we build on this to construct the maximally correlated distributed quantizers which are used in the CorBin-FL and AugCorBin-FL mechanisms in the subsequent sections.
\subsection{Optimality of the LDPQ Stochastic Quantizer}
\begin{algorithm}[t]
\caption{LDPQ: Binary-Output Stochastic Quantization}
\label{Alg:algQ}
\textbf{Procedure:} \textsc{LDPQ}($\epsilon_p, c, r, w$)
\\\textbf{Inputs:}\text{ }Privacy\text{ }budget\text{ }$\epsilon_p$,\text{ }center\text{ }$c$,\text{ }radius\text{ }$r$,\text{ }weight\text{ }$w$
\vspace{-.15in}
\begin{algorithmic}[1]
\STATE $\alpha(\epsilon_p) \gets \frac{e^{\epsilon_p} + 1}{e^{\epsilon_p} - 1}$
\STATE $q \gets \frac{1}{2}+\frac{w-c}{2r\alpha(\epsilon_p)}$\\
\STATE $\!U \!\!\gets \!\!\textsc{BinRand}(q)$ \COMMENT{generates binary U with $P_U(1)=q$}
    \STATE $Q(w) \gets c+Ur\alpha(\epsilon_p)$ 
\RETURN $Q(w)$
\end{algorithmic}
\end{algorithm}
Given $c,r>0$, a
binary-output stochastic quantizer is a (stochastic) mapping $Q:[c\!-\!r,c\!+\!r] \to \{\gamma_1,\gamma_2\}$, where $\gamma_1,\gamma_2\in \mathbb{R}$. 
We quantify privacy in terms of PLDP and utility in terms of the output bias and the MSE. To elaborate, for a given $\epsilon_p>0$, we say the quantizer optimizes the privacy-utility tradeoff if 
 the following conditions are satisfied:
\begin{enumerate}[label=C\arabic*.]
    \item \textbf{Unbiasedness:} $\mathbb{E}(Q(w))\!=\!{w}$ for all $w\in [c\!-\!r,c\!+\!r]$,
    \item \textbf{$\epsilon_p$-PLDP:} $\frac{P(Q(w)=\overline{w})}{P(Q(w')=\overline{w})}\leq e^{\epsilon_p}$ for all $\overline{w}\in \{\gamma_1,\gamma_2\}$ and $w,w'\in [c-r,c+r]$.
    \item \textbf{Minimal MSE:} $\mathbb{E}((Q(w)-w)^2)$ is minimized for all $w\in [c-r,c+r]$. 
\end{enumerate}
The following proposition shows that the $\textsc{LDPQ}$ quantizer uniquely satisfies  conditions C1-C3. 
\begin{Proposition}
\label{prop:1}
Let $c,r,\epsilon_p>0$. The binary-output quantizer $Q(w)= LDPQ(\epsilon_p,c,r,w), w\in [c-r,c+r]$ in Algorithm \ref{Alg:algQ}
    is the unique quantizer satisfying conditions C1-C3. 
\end{Proposition}
The proof is provided in the technical appendix included with the supplementary material.
\subsection{Correlated Stochastic Quantization} 
In the next step, we consider the distributed quantization of a pair of inputs $w,w'$ when the pair of clients have access to a shared sequence of common random bits. To elaborate, let us assume that the clients have access to a sequence of binary symmetric random variables $\mathbf{Z}=(Z_1,Z_2,\cdots,Z_d)$ for a fixed $d\in \mathbb{N}$. We wish to design a pair of stochastic quantizers $Q_i:[c-r,c+r] \times \{0,1\}^d\to \{\gamma_1,\gamma_2\}, i\in \{1,2\}$, satisfying the following conditions:
\begin{enumerate}[label=C\arabic*., start=4]
    \item \textbf{Unbiased Output:} $\mathbb{E}(Q_1(w),\mathbf{Z})=w, \mathbb{E}(Q_2(w'),\mathbf{Z})\!=\!w'$ for all $w,w'\in [c-r,c+r]$.
    \item \textbf{$\epsilon_p$-PLDP:} $\frac{P(Q_i(w,\mathbf{Z})=\overline{w})}{P(Q_i(w',\mathbf{Z})=\overline{w})}\leq e^{\epsilon_p}$ for all $\overline{w}\in \{\gamma_1,\gamma_2\}$, $w,w'\in [c-r,c+r]$, and $i\in \{1,2\}$.
    \item \textbf{Minimal MSE:} $\mathbb{E}((Q_1(w,\mathbf{Z})+Q_2(w',\mathbf{Z})-w-w')^2)$ is minimized for all $w,w'\in [c-r,c+r]$. 
\end{enumerate}
\begin{algorithm}[t]
\caption{CorBinQ: Correlated Stochastic Quantization}
\label{Alg:algCQ}
\textbf{Procedure:}\textsc{CorBinQ}($\epsilon_p, c, r, w, w', d, \mathbf{Z}$)
\\\textbf{Inputs:}\text{ }Privacy\text{ }budget\text{ }$\epsilon_p$,\text{ }center\text{ }$c$,\text{ }radius\text{ }$r$,\text{ }weights $w,w'$, 
\\number\text{ }of\text{ }shared\text{ }random\text{ }bits\text{ }$d$,\text{ }shared\text{ }random\text{ }vector\text{ }$\mathbf{Z}$
\begin{algorithmic}[1]
\STATE $\alpha(\epsilon_p) \gets \frac{e^{\epsilon_p} + 1}{e^{\epsilon_p} - 1}$
\STATE $q_1 \gets \frac{1}{2}+\frac{w-c}{2r\alpha(\epsilon_p)}$, \quad $q_2 \gets \frac{1}{2}+\frac{w'-c}{2r\alpha(\epsilon_p)}$
\STATE $\mathbf{T}_1\gets B_d(\floor{2^dq_1})$,\quad  $\mathbf{T}_2\gets B_d(\floor{2^d(1-q_2)})$
\IF {$\mathbf{Z}\prec \mathbf{T}_1$}
    \STATE $Q_1(w,\mathbf{Z})\gets c+r\alpha(\epsilon_p)$
\ELSIF{$\mathbf{T}_1\prec \mathbf{Z}$}
    \STATE $Q_1(w,\mathbf{Z})\gets c-r\alpha(\epsilon_p)$
\ELSE
    \STATE $U \gets \textsc{BinRand}(2^dq_1-\floor{2^dq_1})$
    \STATE $Q_1(w,\mathbf{Z}) \gets c+Ur\alpha(\epsilon_p)$ 
\ENDIF

\IF {$\mathbf{Z}\prec \mathbf{T}_2$}
    \STATE $Q_2(w',\mathbf{Z})\gets c-r\alpha(\epsilon_p)$
\ELSIF{$\mathbf{T}_1\prec \mathbf{Z}$}
    \STATE $Q_2(w',\mathbf{Z})\gets c+r\alpha(\epsilon_p)$
\ELSE
    \STATE $U' \gets \textsc{BinRand}(2^d(1-q_2)-\floor{2^d(1-q_2)})$
    \STATE $Q_2(w',\mathbf{Z}) \gets c-U'r\alpha(\epsilon_p)$ 
\ENDIF
\RETURN $Q_1(w,\mathbf{Z}),Q_2(w',\mathbf{Z})$
\end{algorithmic}
\end{algorithm}

The pair of stochastic quantizers used in CorBin-FL, denoted by $\textsc{CorBinQ}$ are described in Algorithm \ref{Alg:algCQ}. In the sequel, we show that the pair satisfies conditions C4-C5 for all $d\in \mathbb{N}$, and condition C6 asymptotically as $d\to \infty$. 
To provide an overview, the $\textsc{CorBinQ}$ quantizers described in Algorithm \ref{Alg:algCQ} takes parameters $\epsilon_p,r,c>0$, weights $w,w'\in [c-r,c+r]$, and a shared sequence of independent binary symmetric random variables $\mathbf{Z}$ as input. It then constructs a pair of \textit{threshold vectors} $\mathbf{T}_1$ and $\mathbf{T}_2$ as described in Line 3 of the algorithm, where $B_d(\cdot)$ denotes the d-bit decimal to binary operator. 
The quantization process involves comparing the shared random sequence $\mathbf{Z}$ with these threshold vectors using lexicographic ordering ($\prec$). To elaborate, as described in Lines 4-11 of the algorithm, the first quantizer, $Q_1$, outputs $c+r\alpha(\epsilon_p)$ if $\mathbf{Z} \prec \mathbf{T}_1$ and $c-r\alpha(\epsilon_p)$ if $\mathbf{T}_1 \prec \mathbf{Z}$. Ties are resolved using a locally generated binary random variable $U$ with bias $2^dq_1-\floor{2^dq_1}$, as described in Lines 9-10. On the other hand, as described in Lines 12-19, the second quantizer, $Q_2$, outputs $c-r\alpha(\epsilon_p)$ if $\mathbf{Z} \prec \mathbf{T}_2$ and $c+r\alpha(\epsilon_p)$ if $\mathbf{T}_2 \prec \mathbf{Z}$, with ties resolved by $U'$ as detailed in Lines 17-18. The design of $\mathbf{T}_i$ and the tie-breaking rule ensures that the marginal distribution of the output of each $Q_i$ matches that of the stochastic quantizer used in LDP-FL. Thus, satisfying the unbiasedness and $\epsilon_p$-PLDP conditions. 
Furthermore, $Q_1$ and $Q_2$ produce outputs based on reverse lexicographic ordering relative to each other. This guarantees maximum negative correlation between the quantizers, which minimizes the resulting MSE. 
The following theorem shows that i) the pair of quantizers $(Q_1,Q_2)$ in Algorithm \ref{Alg:algCQ}  satisfy conditions C4-C5 for all $d\in \mathbb{N}$, ii) they optimize the privacy-utility tradeoff described by conditions C4-C6 asymptotically as $d\to \infty$, and iii) provides an upper-bound on the MSE for $d\in \mathbb{N}$. The proof is provided in the technical appendix. 
\begin{Theorem}
\label{th:1}
Let $d\in \mathbb{N}$ and $r,c,\epsilon_p>0$. The pair of correlated stochastic quantizers $(Q_1,Q_2)$ in Algorithm \ref{Alg:algCQ} satisfy conditions C4-C5. Furthermore, let
$\mathcal{Q}$ consist of all pairs of quantizers $(Q'_1,Q'_2)$ satisfying conditions C4-C5, and let the MSE of $(Q'_1,Q'_2)$ for inputs $(w,w')$ be defined as:
    \begin{align*}
 m(Q'_1,Q'_2,w,w')\!=\! \mathbb{E}((Q'_1(w,\mathbf{Z})\!+\!Q'_2(w',\mathbf{Z})\!-\!w\!-\!w')^2),
    \end{align*}
    for all $(Q'_1,Q'_2)\in \mathcal{Q}$.
    Then, the MSE associated with $(Q_1,Q_2)$ is bounded from above as follows:
    \begin{align}
    \label{eq:th:1:opt}
      m(Q_1,Q_2,w,w')\!\!\leq \!\!\!\min_{(Q'_1,Q'_2)\in \mathcal{Q}}\!
  \!\!\!    m(Q'_1,Q'_2,w,w')\!+\!\!\frac{r^2\alpha^2(\epsilon_p)}{2^{d-6}},
    \end{align}
    for all $w,\!w'\!\in\! [r\!-\!c,r\!+\!c]$. Particularly, 
         $m(Q_1,Q_2,w,w')$
         converges uniformly to the minimum MSE over $\mathcal{Q}$ as $d\!\to\! \infty$.
\end{Theorem}

\section{CorBin-FL and AugCorBin-FL Mechanisms}
\begin{algorithm}[t]
\caption{CorBin-FL Mechanism}
\label{Alg:CorBinFL}
\textbf{Procedure:} \textsc{CorBinFL}($n, m, \epsilon_p, \mathbf{c},\mathbf{r}, d,\mathbf{w}_g,(\mathcal{D}_i)_{i\in[n]}$)
\\\textbf{Inputs:} Number of clients $n$, number of model parameters $m$, privacy budget $\epsilon_p$, center vector $\mathbf{c}$, radius vector $\mathbf{r}$, number of shared bits per parameter $d$, global model parameters $\mathbf{w}_g$, distributed datasets $\mathcal{D}_i, i\in [n]$
\begin{algorithmic}[1]
\STATE $\{K_{i,j}\}_{i,j \in [n], i<j} \gets \textsc{DiffieHellmanExchange}(n)$
\STATE $\{(i,p_i)\}_{i\in[n]} \gets \textsc{GenerateRandomPairing}(n)$
\STATE Server shares $\mathbf{w}_g, \mathbf{c}, \mathbf{r}, \{p_i\}_{i\in [n]}, \epsilon_p$ with clients
\STATE Each client $\mathcal{C}_i$ computes update $\mathbf{w}_i$ using dataset $\mathcal{D}_i$
\FOR{each pair of clients $(\mathcal{C}_i, \mathcal{C}_{p_i})$}
    \STATE $Y_i \gets \text{BinRand}(1/2)$ 
    \STATE $Y_{p_i} \gets \text{BinRand}(1/2)$
    \STATE $\textsc{EncryptExchange}(\mathcal{C}_i, \mathcal{C}_{p_i},(Y_i, Y_{p_i}), K_{i,p_i})$
    \IF{$Y_i = Y_{p_i}$}
        \STATE $\ell$ $\gets \mathcal{C}_{\min(i,p_i)}$, $f$ $\gets \mathcal{C}_{\max(i,p_i)}$
    \ELSE
        \STATE $\ell$ $\gets \mathcal{C}_{\max(i,p_i)}$, $f$ $\gets \mathcal{C}_{\min(i,p_i)}$
    \ENDIF
    \FOR{$j \in [m]$}
        \STATE $\mathbf{Z}_{i,j} \gets [\text{BinRand}(1/2)]^d$ \COMMENT{The lead generates sequence of $d$ binary symmetric random variables}
    \ENDFOR
    \STATE $\textsc{EncryptExchange}
    (\ell,f,\{\mathbf{Z}_{i,j}\}_{j=1}^m, K_{i,p_i})$
    \FOR{$j \in [m]$}
        \STATE \!\!\!\!$(\!\overline{W}_{i,j},\! \overline{W}_{p_i,j}) \!\!\gets\!\! \textsc{CorBinQ}(\!\epsilon_p,\! c_j,\! r_j, \! w_{i,j},\! w_{p_i,j},\! d, \!\mathbf{Z}_{i,j}\!)$
    \ENDFOR
    \STATE $\mathcal{C}_i, \mathcal{C}_{p_i}$ send $\overline{\mathbf{W}}_i, \overline{\mathbf{W}}_{p_i}$ to server
\ENDFOR
\STATE $\overline{\mathbf{W}}_g\gets \frac{1}{n}\sum_{i\in [n]}\overline{\mathbf{W}}_i$
\end{algorithmic}
\end{algorithm}

\subsubsection{The CorBin-FL Mechanism:} The mechanism is shown in Algorithm \ref{Alg:CorBinFL}. The clients first establish secure pairwise communications over the public channel facilitated by the server using the Diffie-Hellman key exchange protocol (Line 1) \cite{diffie1976new}. This results in a set of pairwise cryptographic keys $(K_{i,j})_{i,j \in [n], i<j}$. For a detailed discussion on the use of key exchange protocols in FL applications, we refer the reader to \cite{bonawitz2017practical}.
Next, the server generates a random pairing\footnote{A pairing on the set $[n]$ is a partition into disjoint 2-sets.  We denote the pairing by $\{(i,p_i)\}_{i\in [n]}$, where $p_i$ is the pair of $i$.  Note that, by definition, we must have $i=p_{p_i}, i\in [n]$.} on $[n]$, chosen uniformly among all possible pairings (Line 2). We denote this pairing by $\{(i,p_i)\}_{i\in[n]}$, where $p_i$ represents the index of the client paired with $\mathcal{C}_i$.
The server then shares the global model weights $\mathbf{w}_g$, center and radius vectors $\mathbf{c}$ and $\mathbf{r}$, the pairing $\{(i,p_i)\}_{i\in [n]}$, and privacy budget $\epsilon_p$ with the clients (Line 3).
For each pair of clients $(\mathcal{C}_i, \mathcal{C}_{p_i})$, both clients generate binary symmetric random variables $Y_i$ and $Y_{p_i}$, respectively, using $\textsc{BinRand}(1/2)$ (Lines 5-6). They exchange these variables securely using their shared key $K_{i,p_i}$ (Line 8). If $Y_i = Y_{p_i}$, then $\mathcal{C}_{\min(i,p_i)}$ is designated as the lead ($\ell$) and $\mathcal{C}_{\max(i,p_i)}$ as the follow ($f$). The roles are reversed otherwise (Lines 9-13). 
The lead client then generates $m$ sequences of binary symmetric variables $\mathbf{Z}_{i,j} \in \{0,1\}^d$, where $d$ is the number of common random bits shared per model parameter (Lines 14-16). These sequences are sent to the follow client over their encrypted channel (Line 17).
Each pair of clients $(\mathcal{C}_i, \mathcal{C}_{p_i})$ uses the \textsc{CorBinQ} correlated quantization method to generate $(\overline{\mathbf{W}}_i, \overline{\mathbf{W}}_{p_i})$ (Line 19).
The obfuscated updates $(\overline{\mathbf{W}}_i, \overline{\mathbf{W}}_{p_i})$ are then sent to the server (Line 21).
Finally, the server updates the global model weights by averaging the received updates: $\overline{\mathbf{W}}_g \gets \frac{1}{n}\sum_{i\in [n]}\overline{\mathbf{W}}_i$ (line 23).
Note that if $d=0$, then the CorBin-FL mechanism reduces to the LDP-FL mechanism.

\subsubsection{Privacy-Utility Guarantees:} From Theorem \ref{th:1} it follows that CorBin-FL achieves  $\epsilon_p$-PLDP.
The following theorem states the utility guarantees of CorBin-FL, in terms of unbiasedness and achievable MSE on the server-side.
\begin{Theorem}
\label{prop:util}
    The CorBin-FL mechanism is unbiased, i.e., 
$        \mathbb{E}(\overline{\mathbf{W}}_g) =\frac{1}{n}\sum_{i\in [n]} \mathbf{w}_i$,
    where $\mathbf{w}_i, i\in [n]$ are the local client updates, and $\overline{\mathbf{W}}_g$ is the average of the obfuscated updates at the server. Furthermore, the MSE is bounded by:
    \begin{align*}
       & \lim_{d\to \infty}\mathbb{E}((\overline{{W}}_{g,j}-\frac{1}{n}\sum_{i\in [n]} {w}_{i,j})^2)
       \\&\leq\frac{r_j^2}{2n}\left(\!(\sqrt{2}\!-\!1)\alpha(\epsilon_p)\!+\!1\right)\left(\! (\sqrt{2}\!+\!1\!)\alpha(\epsilon_p)\!-\!1\!\right),\quad j\in [m]
    \end{align*}
\end{Theorem}
The proof is provided in the technical appendix. As shown in \cite[Lemma 3]{sun2020ldp}, the mean square error under LDP-FL is close to $\frac{r^2\alpha^2(\epsilon_p)}{n}$ for large $n$. Consequently, for small values of $\epsilon_p$, where $\alpha(\epsilon_p)\gg 1$ and large values of $n$, the CorBin-FL mechanism improves the mean-square error by at least a factor of two compared to LDP-FL.

\subsubsection{The Augmented CorBin-FL Mechanism:}
So far, we have shown that CorBin-FL is unbiased, achieves $\epsilon_p$-PLDP, and improves upon LDP-FL in terms of MSE. In addition to PLDP, the CorBin-FL mechanism can be modified to achieve UCDP and SCDP. To elaborate, let $\gamma\in [0,1]$ and consider a mechanism where, at each round, $\gamma$ fraction of the clients use the LDP-FL mechanism to obfuscate their corresponding weights, and $(1-\gamma)$ fraction use the CorBin-FL mechanism. We call the resulting mechanism, which is a hybrid mechanism between LDP-FL and CorBin-FL, the AugCorBin-FL mechanism.
The following theorem prides the UCDP guarantees and an upper-bound on the mean square error resulting from the AugCorBin-FL mechanism.

\begin{Theorem}
\label{th:3}
Let $\epsilon_p,\delta>0$, and $\gamma\in (0,1]$. Let us define $r=\max_{j\in [m]}r_j$ and assume that $n$ is large enough that
\begin{align*}
&(n\gamma-1)(\frac{1}{4}-\frac{1}{4\alpha^2(\epsilon_p)})\geq \max(23\log{\frac{m}{\delta}}, 2r\alpha(\epsilon_p)).
\end{align*}
Then, the AugCorBin-FL mechanism achieves $(\epsilon_u,\delta)$-UCDP and $\epsilon_p$-PLDP, where 
\begin{align}
&\label{eq:th:3:1}
\frac{\epsilon_u}{r\alpha(\epsilon_p)}= 
\sqrt{\frac{8m\log{\frac{1.25}{\delta}}}{(n\gamma-1)e_p}}+\frac{8(\log{\frac{1.25}{\delta}+\log{\frac{20m}{\delta}\log{\frac{10}{\delta})}}}}{3(n\gamma-1)}
+\frac{4b_p\sqrt{2m}(1.75+\frac{3.75}{\alpha^2(\epsilon_p)})\sqrt{\log{\frac{10}{\delta}}}}{(n\gamma-1)(1-\frac{\delta}{10})e_p},\\
&e_p= (1+\frac{1}{\alpha^2(\epsilon_p)}),\quad 
b_p= \frac{1}{3}e_p+\frac{1}{\alpha(\epsilon_p)}.\nonumber
\end{align}
Furthermore, the mean square error is bounded by:
    \begin{align*}
       & \lim_{d\to \infty}\mathbb{E}((\overline{{W}}_{g,j}-\frac{1}{n}\sum_{i\in [n]} {w}_{i,j})^2)
       \leq \frac{\gamma r_j^2\alpha^2(\epsilon_p)}{n}+
       \frac{(1-\gamma)}{2n}r_j^2\left((\sqrt{2}\!-\!1)\alpha(\epsilon_p)+1\right)\left(  (\sqrt{2}+1)\alpha(\epsilon_p)\!-\!1\right),
    \end{align*}
    for all $j\in [m]$.
\end{Theorem}
The proof is provided in the technical appendix in the supplementary material. The SCDP guarantees for AugCorBin-FL can be derived in terms of the sensitivity parameter, using a similar method as in the proof of Theorem \ref{th:3}.

\subsubsection{Robustness to Client Dropout:} The CorBin-FL mechanism is robust to client dropouts. The reason is that if a member of a pair of
clients drops out, the quantization performed by the other client is equivalent to the LDPQ quantizer. In fact, if a lower-bound to the dropout probability of each individual client is known, and if the clients dropout independently of each other, then one can derive additional central differential privacy guarantees for the resulting mechanism. This is made precise in the following theorem.

\begin{Theorem}
\label{prop:4}
Let $\epsilon_p,\delta,p,t>0$, and $\gamma\in (0,1]$.  Assume that clients drop out independently of each other with dropout probability $p$ in each communication round. Furthermore, let us define $r=\max_{j\in [m]}r_j$ and assume that $n$ is large enough and there exists $\delta_1<\delta$ such that
\begin{align*}
&(n\gamma-1)(\frac{1}{4}-\frac{1}{4\alpha^2(\epsilon_p)})\geq \max(23\log{\frac{m}{\delta_1}}, 2r\alpha(\epsilon_p)),
\\& \gamma =p(1-p)-\sqrt{\frac{p(1-p)(1-2p(1-p))}{4(\delta-\delta_1)n}}.
\end{align*}
Then, the CorBin-FL mechanism achieves $\epsilon_p$-PLDP and $(\epsilon_u,\delta)$-UCDP, where, $\epsilon_u$ is given  by Equation \eqref{eq:th:3:1}. Furthermore, the mean square error is bounded by: 
    \begin{align*}
       & \lim_{d\to \infty}\mathbb{E}((\overline{{W}}_{g,j}\!-\!\frac{1}{N}\!\!\sum_{i:F_i=1}\!\! {w}_{i,j})^2)\!\leq \!\frac{\gamma' r^2\alpha^2(\epsilon_p)}{n\theta}\!+\! 8r^2\alpha^2(\epsilon_p)\delta_1
     \\&+
       \frac{(1-\gamma)}{2n\theta}r^2\left((\sqrt{2}-1)\alpha(\epsilon_p)+1\right)\left(  (\sqrt{2}+1)\alpha(\epsilon_p)-1\right).
    \end{align*}
    for all $j\in [m]$, where $F_i$ is the indicator that the $i$th client does not drop out, $N= \sum_{i\in [n]}F_i$, and 
        \begin{align*}
        &\gamma'= p(1-p)+\sqrt{\frac{p(1-p)(1-2p(1-p))}{4(\delta-\delta_1)n}},
        \\&\theta=p(1-p)-\sqrt{\frac{p(1-p)}{(\delta-\delta_1)n}}.
    \end{align*}
\end{Theorem}
The proof is provided in the technical appendix.
\subsubsection{Communication Cost:} 
The server broadcasts the global model to all clients, which requires $\mathcal{O}(32m)$ bits of downlink communication, assuming 32-bit floating-point representation. 
Subsequently, the DH key exchange protocol is used to enable pairwise secure communication among clients. The DH protocol has uplink and downlink communication cost of $\mathcal{O}(nk)$ bits per client, where $k$ represents the DH key size \cite{diffie1976new}. 
Then, the leaders in each pair of clients 
generate a $d$-length binary symmetric random vector, and encrypt and transmit the vector to the follow, which requires $\mathcal{O}(dmk)$ bits of uplink and downlink communication. Each client applies \textsc{CorBinQ} and uploads the quantized updated model to the server, with $\mathcal{O}(m)$ bits of uplink communication. The aggregate communication complexity per round is  $\mathcal{O}(dmk + nk)$. Typically, the number of model parameters $m$ dominates the number of clients $n$. Consequently, the communication cost is $\mathcal{O}(dmk)$.

\noindent\textbf{Computation Cost:} The computation cost of the DH key exchange protocol is $\mathcal{O}(nk^2 \log k)$. For generating $dm$ binary symmetric random variables, the cost is $\mathcal{O}(dm)$, and the encryption and computational complexity of CorBinQ in Algorithm \ref{Alg:algCQ} consisting of a lexicographic binary vector comparison and the generation of a binary random variable is $\mathcal{O}(m)$ per client. In total the computation complexity is  $\mathcal{O}(dm+nk^2 \log k$).


\section{Empirical Analysis}
\begin{figure}[t]
\centering
\includegraphics[width=0.6\columnwidth]{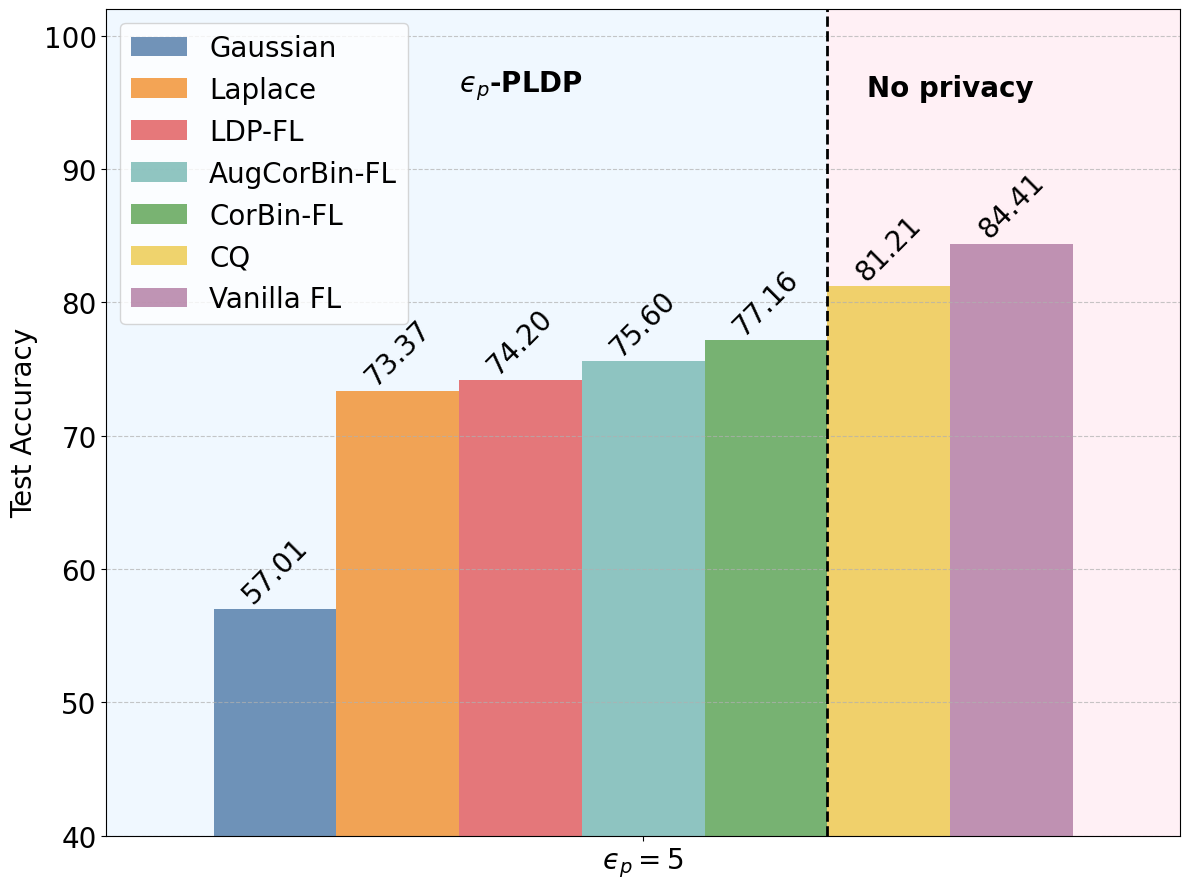} 
\vspace{-.1in}
\caption{Comparison of different privacy mechanisms (Experiment 1).}
\vspace{-.2in}
\label{fig: all}
\end{figure} 
\subsubsection{Experimental Setup:}
\begin{figure*}[b]
    \centering
        \begin{subfigure}[b]{0.32\textwidth}
        \centering
        \includegraphics[width=\textwidth]{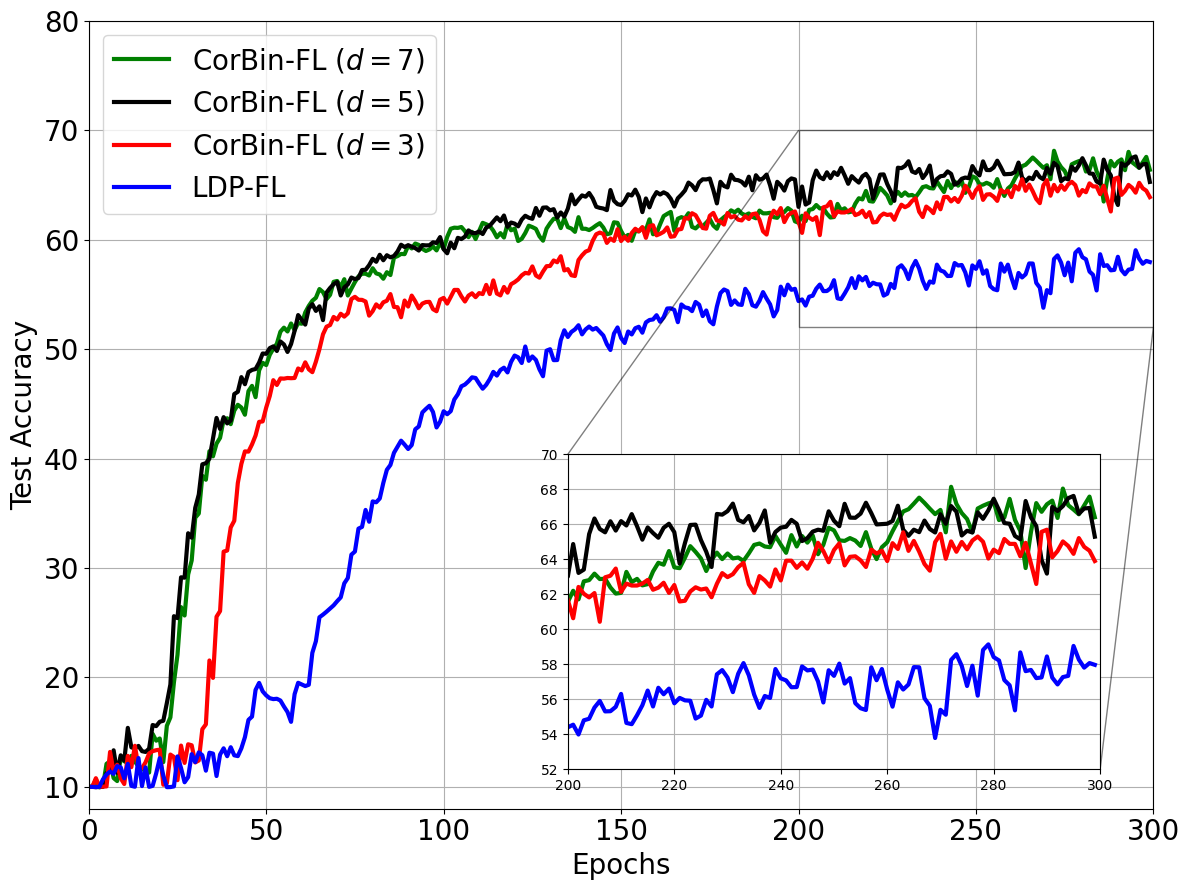}
        \caption{}
        \label{fig: NumRand_EP1}
    \end{subfigure}
    \hfill
        \begin{subfigure}[b]{0.33\textwidth}
        \centering
 \centering
\includegraphics[width=\textwidth]{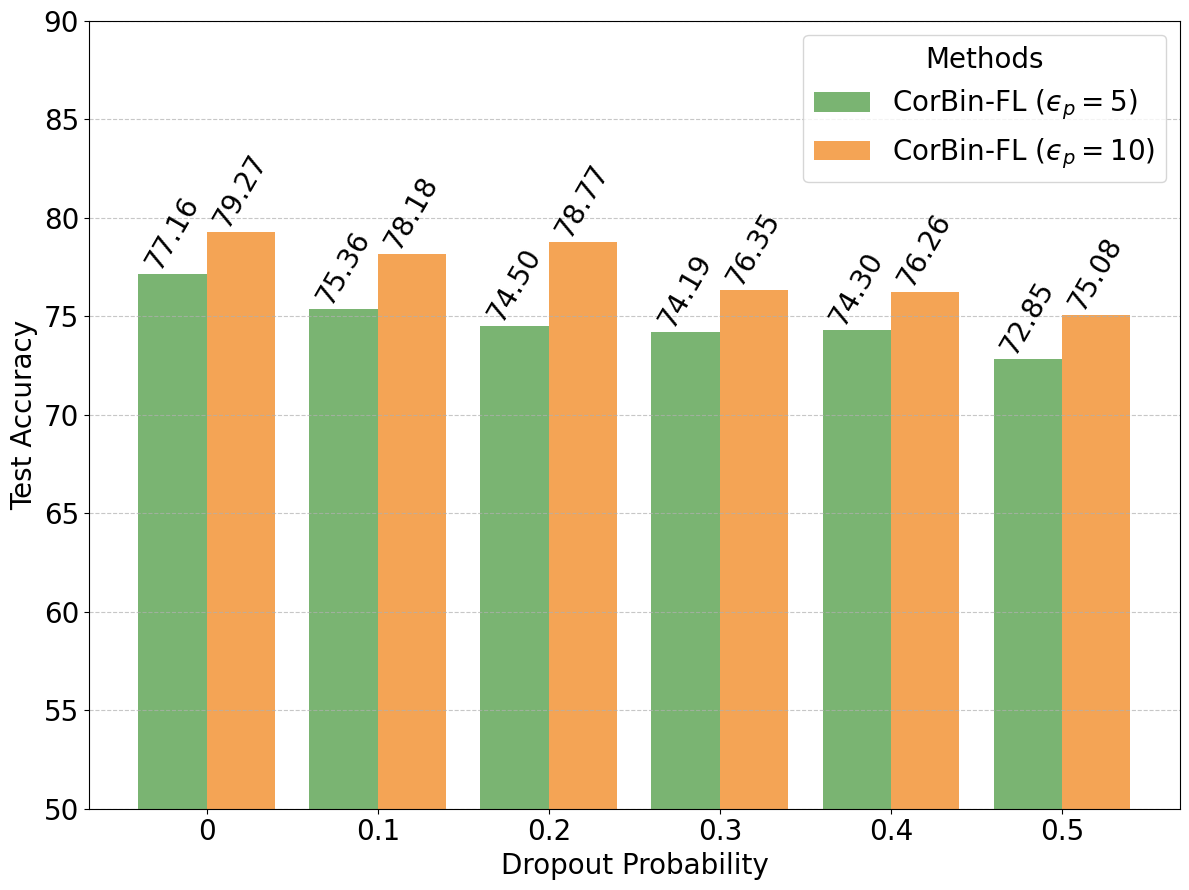} 
\caption{}
\label{fig: dropout}
 
    \end{subfigure}
    \hfill
    \begin{subfigure}[b]{0.32\textwidth}
        \centering
        \includegraphics[width=\textwidth]{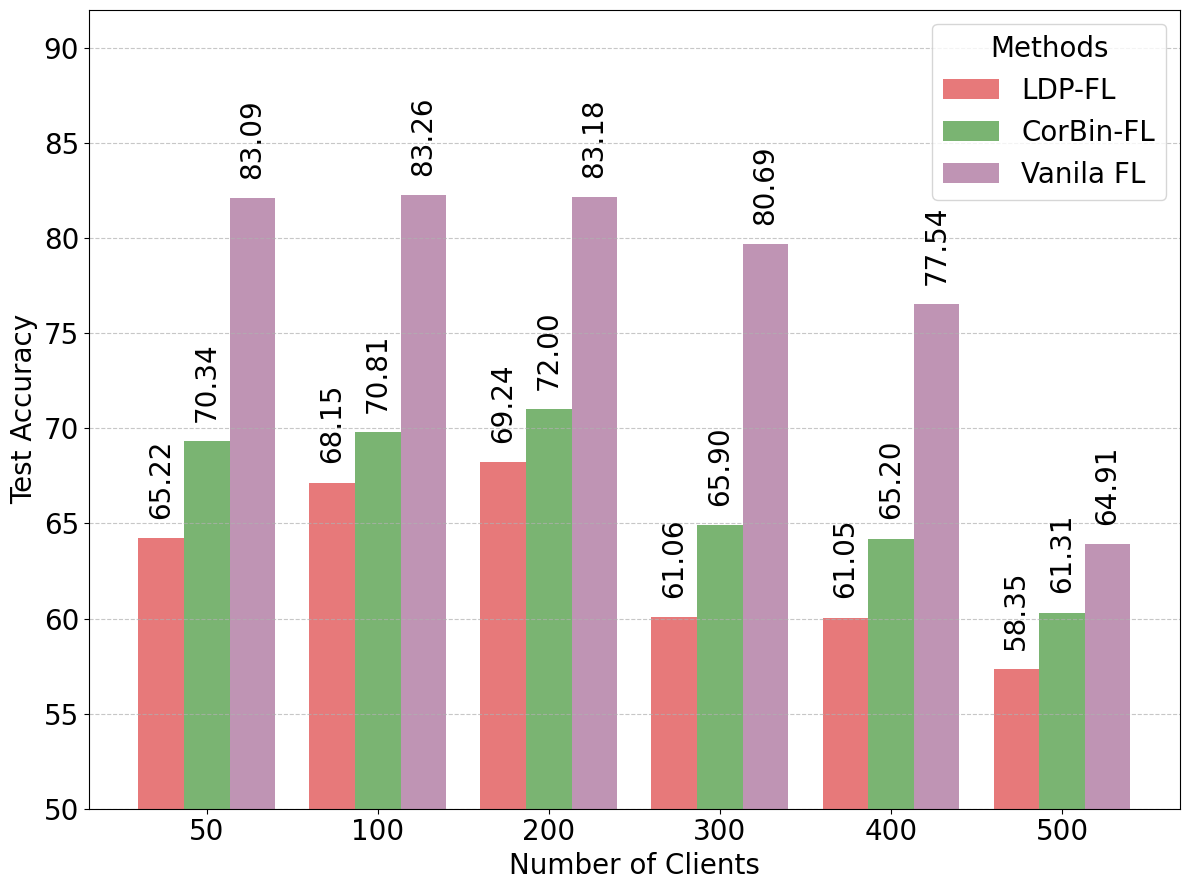}
        \caption{}
        \label{fig: clients}
    \end{subfigure}
    \caption{Experimental results for (a) number of shared common random bits (Experiment 2), (b) dropout probability (Experiment 3) and (c) number of clients (Experiment 4).}
    \vspace{-.05in}
    \label{fig:all_comparisons}
\end{figure*}
We empirically evaluate the performance of the proposed privacy mechanisms over a series of experiments as outlined in the sequel. To provide an overview, our objective is to demonstrate that i) under a fixed privacy budget and number of communication rounds, the proposed mechanisms outperform the LDP-FL, and Gaussian and Laplace Mechanisms in terms of overall model accuracy (Figure \ref{fig: all}), ii) the accuracy gains increase as the number of shared random bits per model parameter is increased, however, the gains saturate for relatively small values of $d$ (e.g., $d= 5$) as predicted in Theorem \ref{th:1} (Figure \ref{fig: NumRand_EP1}), iii) the proposed methods are robust to client dropouts, and the model accuracy does not fall significantly even for large dropout probabilities, e.g., for $p= 0.5$ (Figure \ref{fig: dropout}), and iv) the proposed methods outperform the LDP-FL mechanism for different numbers of clients ranging from 50 to 500 (Figure \ref{fig: clients}). A detailed description of the experimental setup is provided in this section. Further empirical evaluations and ablation studies are included in the technical appendix. 

\subsubsection{Datasets and Models:} Experiments are over the MNIST \cite{lecun2010mnist} and CIFAR10 \cite{krizhevsky2009learning} datasets. The samples are split randomly, with $80\%$ of the samples used for training and the remaining samples for testing. We choose 20\% of the training samples randomly for validation.
We consider an IID data distribution setting, where each client receives an equal fraction of the samples randomly and without replacement. The batch size is 64 for all experiments except for Figure \ref{fig: clients}, where it is taken to be 32 due to the large numbers of clients which yields a smaller per-client dataset size.  We use the ResNet18 \cite{he2016deep} model for experiments over the CIFAR10 dataset and small (two-layers) VGG \cite{simonyan2014very} model for experiments over the MNIST dataset. The number of shared random bits per parameter is $d=5$ and the AugCorBin-FL parameter is taken as $\gamma=0.2$ across all experiments unless specified otherwise. The center and radius are shared by the server at each round based on the layer-wise minimum and maximum  parameter values in the previous round. The experiments are performed on a single  NVIDIA A100-PCIE-40GB GPU.
\subsubsection{Smoothed Update Rule and Checkpoints:} We include a global learning rate hyperparatmer $\lambda$. At each communication round, let $\mathbf{w}_{g,init}$ be the initial global parameter vector, and $\mathbf{w}_{g,final}$ the updated parameter vector at the end of the communication round. We update the model as follows:
\begin{equation}
    \mathbf{w}_g = (1-\lambda)\mathbf{w}_{g,init} +\lambda \mathbf{w}_{g,final}
\end{equation}
Taking $\lambda=1$ reduces to the original update rule without the inclusion of the global learning rate hyperparameter.  For each experiment, and for all privacy mechanisms, we preform a grid search over $\lambda\in\{0.1,0.2,...,1\}$. We also implement a check-pointing mechanism with a patience of five epochs, so that the model is reloaded if the validation accuracy does not increase over five epochs. 
\subsubsection{Experiment 1 - Comparison of Privacy Mechanisms:} We train an FL model over CIFAR-10 for 50 clients, with a PLDP budget of $\epsilon_p=5$ using the Gaussian, Laplacian, LPD-FL, CorBin-FL, and AugCorBin-FL mechanisms. 
For the Gaussian mechanism, we find the noise variance using the bound in \cite{balle2018improving}. The Laplace mechanism is implemented according to \cite{dwork2006calibrating}. Figure \ref{fig: all} shows the resulting accuracy gains. To provide further baselines for comparison, we have implemented the vanilla-FL with no privacy guarantees. An advantage of CorBin-FL and AugCorBin-FL over other mechanisms is the low communication cost due to one-bit quantization of the model parameters. Thus, to provide another baseline, we have implemented the correlated quantization (CQ) method of \cite{suresh2022correlated} which does not provide piracy guarantees but reduces the communication cost via quantization. 
\subsubsection{Experiment 2 - Effects of the Number of Shared Random Bits} 
Under the same settings as the previous experiment and with $\epsilon_p=1$, we evaluate the performance of CorBin-FL under various number of shared random bits per model paramter, $d\in \{3,5,7\}$. The results are shown in Figure \ref{fig:all_comparisons}(a). 

\subsubsection{Experiment 3 - Effects of Dropout:} Under the same settings as the previous experiment, we have evaluated the effect of random client dropout on the performance of CorBin-FL. As shown in Figure \ref{fig:all_comparisons}(b), the performance does not fall significantly even when the client randomly and independently dropout at each round with $50\%$ probability. 

\subsubsection{Experiment 4 - Effect of the Number of Clients:} We have evaluated the performance of LDP-FL, CorBin-FL with privacy budget $\epsilon_p = 5$, and Vanilla-FL without privacy guarantees. The results are shown in Figure \ref{fig:all_comparisons}(c). It can be observed that CorBin-FL consistently outperforms LDP-FL given a fixed privacy budget over a range of client numbers.

\section{Conclusion and Future Works}
The CorBin-FL and AugCorBin-FL mechanisms were introduced for differentially private federated learning. These approaches, which are based on correlated binary stochastic quantization, were shown to achieve various notions of differential privacy. Theoretical guarantees for privacy parameters and mean squared error were derived. Empirical evaluations on MNIST and CIFAR10 datasets demonstrated improved model accuracy compared to existing methods under equal PLDP privacy budgets. In future works, we will investigate the extension of the proposed methods to correlated quantizers with larger output alphabets, and correlated quantization among collections of more than two clients. 
\section{Acknowledgement}
This work is based upon the work partly  supported by the National Center for Transportation Cybersecurity and Resiliency (TraCR) (a U.S. Department of Transportation National University Transportation Center) headquartered at Clemson University, Clemson, South Carolina, USA. Any opinions, findings, conclusions, and recommendations expressed in this material are those of the author(s) and do not necessarily reflect the views of TraCR, and the U.S. Government assumes no liability for the contents or use thereof.
\clearpage
\newpage

\IEEEtriggeratref{41}
\bibliographystyle{IEEEtran}
\bibliography{aaai25}
\clearpage

\begin{appendices}

\section{Proof of Proposition \ref{prop:1}}
\label{App:prop:1}
   For $w\in [c-r,c+r]$, let $p_w= P(Q(w)=\gamma_1)$.
    From condition C1, we have:
    \begin{align}
        \label{eq:prop:1:0}
        p_w\gamma_1+(1-p_w)\gamma_2=w \Rightarrow p_w= \frac{\gamma_2-w}{\gamma_2-\gamma_1}.
    \end{align}
    From condition C2, for $w,w'\in [c-r,c+r]$ and $\overline{w}= \gamma_1$, we have:
    \begin{align*}
    \frac{P(Q(w)=\overline{w})}{P(Q(w')=\overline{w})}\leq e^{\epsilon_p}\Rightarrow     \frac{\frac{\gamma_2-w}{\gamma_2-\gamma_1}}{\frac{\gamma_2-w'}{\gamma_2-\gamma_1}}=\frac{\gamma_2-w}{\gamma_2-w'}\leq e^{\epsilon_p}.
    \end{align*}
    To ensure that the condition holds for all $w,w' \in [c-r,c+r]$, 
    it suffices to ensure the condition holds when the numerator is maximized and the denominator is minimized, i.e., for $w=c-r$ and $w'=c+r$. We have:
    \begin{align}
        \label{cst:1}
        \frac{\gamma_2-c+r}{\gamma_2-c-r}\leq e^{\epsilon_p}
        \Rightarrow \gamma_2\geq c+r\left(\frac{e^{\epsilon_p}+1}{e^{\epsilon_p}-1}\right)= c+r\alpha(\epsilon_p).
    \end{align}
    Similarly, by taking $\overline{w}=\gamma_2$, we have:
    \begin{align}
    \label{cst:2}
        \gamma_1\leq c-r\left(\frac{e^{\epsilon_p}+1}{e^{\epsilon_p}-1}\right)=c-r\alpha(\epsilon_p).
    \end{align}
    Note that
    \begin{align*}
        &\mathbb{E}((Q(w)-w)^2)= \mathbb{E}(Q^2(w))-2w\mathbb{E}(Q(w))+w^2
        \\&= \mathbb{E}(Q^2(w))-w^2,
    \end{align*}
    where we have used the unbiasedness property in the last equality. So, Condition C3 requires the minimization of the following:
    \begin{align*}
        \mathbb{E}(Q^2(w))&= p_w \gamma_1^2+(1-p_w)\gamma_2^2
        \\& = \left(\frac{\gamma_2-w}{\gamma_2-\gamma_1}\right)\gamma_1^2+\left(\frac{w-\gamma_1}{\gamma_2-\gamma_1}\right)\gamma_2^2
        \\&= \frac{-\gamma_1\gamma_2(\gamma_2-\gamma_1)+(\gamma_2-\gamma_1)(\gamma_2+\gamma_1)w}{\gamma_2-\gamma_1}
        \\&= -\gamma_1\gamma_2+(\gamma_1+\gamma_2)w.
    \end{align*}
    Note that since $w\in [c-r,c+r]\subseteq [\gamma_1,\gamma_2]$ from 
    Equations \eqref{cst:1} and \eqref{cst:2}, the derivative of the above term is positive with respect to $\gamma_2$ for all values of $\gamma_1$, and it is  negative with respect to $\gamma_1$ for all values of $\gamma_2$. Thus, from the
    constraints of Equations \eqref{cst:1} and \eqref{cst:2}, we get $\gamma_1=c-r\alpha(\epsilon_p)$ and $\gamma_2=c+r\alpha(\epsilon_p)$. It remains to show that $P(Q(w)=c-r\alpha(\epsilon_p))= \frac{1}{2}-\frac{w-c}{2r\alpha(\epsilon_p)}$, which follows from Equation \eqref{eq:prop:1:0}: 
    \[p_w=\frac{c+r\alpha(\epsilon_p)-w}{c+r\alpha(\epsilon_p)-c+r\alpha(\epsilon_p)}= \frac{1}{2}-\frac{w-c}{2r\alpha(\epsilon_p)}.\]
    This completes the proof. 
\qed
\section{Proof of Theorem \ref{th:1}}
To verify that $(Q_1,Q_2)$ satisfy condition C4, first, we show that the marginal distribution of each of the quantizers is equal to that of the corresponding LDP-FL quantizer:
\begin{align*}
    &P\left(Q_1(w,\mathbf{Z})= c-r\alpha(\epsilon_p)\right) =P(\mathbf{T}_1\prec \mathbf{Z} )+P({\mathbf{T}}_1= \mathbf{Z})P(U=-1)
    \\&= 1-\frac{1}{2^{d}} \floor{2^d q_1}-\frac{1}{2^d}+ \frac{1}{2^d}(1-2^dq_1+ \floor{2^dq_1}) 
    = 1-q_1=\frac{1}{2}-\frac{w-c}{2r\alpha(\epsilon_p)}.
\end{align*}
Similarly, 
\begin{align*}
   &P\left(Q_2(w',\mathbf{Z})= c-r\alpha(\epsilon_p)\right) =  \frac{1}{2}-\frac{w'-c}{2r\alpha(\epsilon_p)}.
\end{align*}
So, following the arguments in the proof of Proposition \ref{prop:1}, conditions C4 and C5 are satisfied by $(Q_1,Q_2)$.

To prove  the bound in Equation \eqref{eq:th:1:opt}, let us take an arbitrary pair of stochastic quantizers $(Q_1^*,Q_2^*)\in \mathcal{Q}$, and define 
    \begin{align*}
    &P^*_{i,j}=P(Q^*_1(w,\mathbf{Z})=\gamma_i, Q^*_2(w'
,\mathbf{Z})=\gamma_j),
    \end{align*}
where $i,j\in \{1,2\}$. Following the arguments in the proof of Proposition \ref{prop:1}, from condition C4, we have:
    \begin{align}
    \nonumber
    &w=\mathbb{E}(Q^*_1(w,\mathbf{Z}))= 
    P(Q^*_1(w,\mathbf{Z})=\gamma_1)
    \gamma_1+  (1- P(Q^*_1(w,\mathbf{Z})=\gamma_1))\gamma_2
    \\&
    \Rightarrow 
     P(Q^*_1(w,\mathbf{Z})=\gamma_1)=
    \frac{\gamma_2-w}{\gamma_2-\gamma_1}
    \nonumber\\&
       \label{eq:th:1:1}
\Rightarrow   P^*_{1,1}+P^*_{1,2}=
    \frac{\gamma_2-w}{\gamma_2-\gamma_1}.
    \end{align}
    Note that by definition:
    \begin{align}
    \label{eq:th:1:1.5}
        P^*_{1,1}+P^*_{1,2}+P^*_{2,1}+P^*_{2,2}= 1.
    \end{align}

    Similar to Equation \eqref{eq:th:1:2}, from $w'=\mathbb{E}(Q^*_2(w',\mathbf{Z}))$ in condition C4, we have: 
    \begin{align}
    \label{eq:th:1:2}
        P^*_{1,1}+P^*_{2,1}= \frac{\gamma_2-w'}{\gamma_2-\gamma_1}
    \end{align}

Furthermore, from condition C5, we have:
\begin{align}
 \nonumber&\frac{P(Q^*_i(w,\mathbf{Z})=\gamma_1)}{P(Q^*_i(w',\mathbf{Z})=\gamma_1)}
    = \frac{\gamma_2-c-r}{\gamma_2-c+r}\leq e^{\epsilon_p}
     \\&\label{eq:th:1:3}    \Rightarrow \gamma_2\geq c+r-\frac{2re^{\epsilon_p}}{e^{\epsilon_p}-1}= c+r\alpha(\epsilon_p).
\end{align}
Similarly,    
\begin{align}
\label{eq:th:1:4}
    \gamma_1\leq c-r\alpha(\epsilon_p)
\end{align} 
Lastly, from condition C6, the following should be minimized for all $w,w'\in [c-r,c+r]$:
\begin{align*}
     &  \mathbb{E}(({Q^*_1}(w,\mathbf{Z})+{Q^*_2}(w',\mathbf{Z})-w-w')^2)
     \\& = 
       \mathbb{E}(({Q^*_1}(w,\mathbf{Z})-c+{Q^*_2}(w',\mathbf{Z})-
       c
       -(w-c)-(w'-c))^2)
\\& =  \mathbb{E}(({Q^*_1}(w,\mathbf{Z})-c+{Q^*_2}(w',\mathbf{Z})-
       c)^2)
       -(w+w'-2c)^2,
\end{align*}
where in the last equality, we have used the unbiasedness property. Consequently, we need to minimize:
\begin{align*}
    &\mathbb{E}(({Q^*_1}(w,\mathbf{Z})-c)^2)+\mathbb{E}(({Q^*_2}(w',\mathbf{Z})-c)^2)+
    \\&
    2\mathbb{E}(({Q^*_1}(w,\mathbf{Z})-c)({Q^*_2}(w',\mathbf{Z})-c)).
\end{align*}
    Note that:
    \begin{align}
        &\mathbb{E}(({Q^*_1}(w,\mathbf{Z})-c)^2)
        = (P^*_{1,1}+P^*_{2,1})(\gamma_1-c)^2+ (P^*_{1,2}+P^*_{2,2})(\gamma_2-c)^2.
    \end{align}
  
Similarly, 
    \begin{align}
          &\mathbb{E}((Q^*_2(w',\mathbf{Z})-c)^2)
        = (P^*_{1,1}+P^*_{1,2})(\gamma_1-c)^2+ (P^*_{2,1}+P^*_{2,2})(\gamma_2-c)^2.
    \end{align}
    Furthermore,
    \begin{align}
         \nonumber &\mathbb{E}((Q^*_1(w,\mathbf{Z})-c)(Q^*_2(w',\mathbf{Z})-c))
 \\&\nonumber = P^*_{1,1}(\gamma_1-c)^2+
(P^*_{1,2}+P^*_{2,1})(\gamma_1-c)(\gamma_2-c)
       +
        P^*_{2,2}(\gamma_2-c)^2,
    \end{align}
    Consequently, we have:
    \begin{align*}
        &  \mathbb{E}((Q^*_1(w,\mathbf{Z})-c)^2)+\mathbb{E}((Q^*_2(w',\mathbf{Z})-c)^2)
     +2\mathbb{E}(({Q^*_1}(w,\mathbf{Z})-c)({Q^*_2}(w',\mathbf{Z})-c))
      \\&= P^*_{1,1}(4(\gamma_1-c)^2)
      +P^*_{2,2}(4(\gamma_2-c)^2)
      +(P^*_{1,2}+P^*_{2,1})(\gamma_1-c+\gamma_2-c)^2.
    \end{align*}
Let us define $\tilde{w}=w-c$, $\tilde{w}'= w'-c$, $\tilde{\gamma_1}=\gamma_1-c$, and $\tilde{\gamma_2}=\gamma_2-c$. Then the optimization problem can be written as the minimization of:
\begin{align}
\label{eq:th:1:opt:tilde}
    P^*_{1,1}(4\tilde{\gamma}_1^2)
      +P^*_{2,2}(4\tilde{\gamma}_2^2)
      +(P^*_{1,2}+P^*_{2,1})(\tilde{\gamma}_1\tilde{\gamma}_2)^2.
\end{align}
Furthermore, the constraints in
Equations \eqref{eq:th:1:1} and \eqref{eq:th:1:2} can be rewritten as:
\begin{align*}
   & P^*_{1,1}+P^*_{1,2}=
    \frac{\tilde{\gamma}_2-\tilde{w}}{\tilde{\gamma}_2-\tilde{\gamma}_1},
    \\&
   P^*_{1,1}+P^*_{2,1}=
    \frac{\tilde{\gamma}_2-\tilde{w}'}{\tilde{\gamma}_2-\tilde{\gamma}_1}.
\end{align*}
We prove the theorem for the case where $|\tilde{\gamma}_1|\leq |\tilde{\gamma}_2|$. The proof for the case where $|\tilde{\gamma}_1|> |\tilde{\gamma}_2|$ follows by symmetry.
We consider two subcases:
\\\textbf{Case 1:} $|\tilde{\gamma_1}|\leq |\tilde{\gamma_2}|$ and $\tilde{w}+\tilde{w}'\leq \tilde{\gamma}_1+\tilde{\gamma}_2$:
\\Note that in this case, we have:
\begin{align*}
     \frac{\tilde{\gamma}_2-\tilde{w}}{\tilde{\gamma}_2-\tilde{\gamma}_1}+ \frac{\tilde{\gamma}_2-\tilde{w}'}{\tilde{\gamma}_2-\tilde{\gamma}_1}\geq 1.
\end{align*}
Furthermore, by the assumptions of Case 1,  we have $4\tilde{\gamma}_1^2\leq 4\tilde{\gamma}_2^2$. So, Equation \eqref{eq:th:1:opt:tilde} is minimized by taking:
\begin{align}
\label{eq:th:1:JD:1}
    &P^*_{1,1}= \frac{\tilde{\gamma}_2-\tilde{w}}{\tilde{\gamma}_2-\tilde{\gamma}_1}+ \frac{\tilde{\gamma}_2-\tilde{w}'}{\tilde{\gamma}_2-\tilde{\gamma}_1}-1,
    \quad P^*_{1,2}
    = 1-  \frac{\tilde{\gamma}_2-\tilde{w}'}{\tilde{\gamma}_2-\tilde{\gamma}_1},
    \\& \label{eq:th:1:JD:2}
    P^*_{2,1}
    = 1-  \frac{\tilde{\gamma}_2-\tilde{w}}{\tilde{\gamma}_2-\tilde{\gamma}_1},
    \qquad P^*_{2,2}=0.
\end{align}
It should be noted that there is no guarantee that there exists a pair of stochastic quantizers $(Q^*_1,Q^*_2)$ achieving the above distribution. Thus, minimizing the mean square error over $\tilde{\gamma}_1,
\tilde{\gamma}_2$ by considering the above distribution only yields a lower bound on the achievable mean square error. 
Thus, we have:
\begin{align}
    \nonumber&
    m(Q^*_1,Q^*_2,w,w')\geq \min_{\tilde{\gamma}_1,\tilde{\gamma}_2}
    \frac{\tilde{\gamma}_1+\tilde{\gamma}_2-\tilde{w}-\tilde{w}'}{\tilde{\gamma}_2-\tilde{\gamma}_1}(4\tilde{\gamma}^2_1)
\\& \nonumber   +\frac{-2\tilde{\gamma}_1+\tilde{w}+\tilde{w}'}{\tilde{\gamma}_2-\tilde{\gamma}_1}(\tilde{\gamma}_1+\tilde{\gamma}_2)^2
    \\&\nonumber
    =\min_{\tilde{\gamma}_1,\tilde{\gamma}_2} \frac{2\tilde{\gamma}_1^3\!-\!2\tilde{\gamma}_1\tilde{\gamma}_2^2\!
    +(\tilde{w}\!+\!\tilde{w}')(
-3\tilde{\gamma}_1^2+\!2\tilde{\gamma}_1\tilde{\gamma}_2+\!\tilde{\gamma}_2^2)}{\tilde{\gamma}_2\!-\!\tilde{\gamma}_1}
    \\
    \label{eq:th:1:6}&    =\min_{\tilde{\gamma}_1,\tilde{\gamma}_2}-2\tilde{\gamma}_1(\tilde{\gamma}_1+\tilde{\gamma}_2)+(\tilde{w}+\tilde{w}')(3\tilde{\gamma}_1+\tilde{\gamma}_2).
\end{align}
Recall that $\tilde{\gamma}_1\leq -r\alpha(\epsilon_p)\leq -r$, hence the derivative of Equation \eqref{eq:th:1:6} with respect to $\tilde{\gamma}_2$ is always non-negative since $2\tilde{\gamma}_1\leq \tilde{w}+\tilde{w}'$ for all $w,w'\in [c-r,c+r]$. So, the $\tilde{\gamma}_2$ is optimized by taking $|\tilde{\gamma}_2|=|\tilde{\gamma}_1|$ (which is the minimum value considering the conditions assumed in Case 1).  Thus, Equation \eqref{eq:th:1:6} simplifies to:
\begin{align*}
\min_{\tilde{\gamma}_1}2(\tilde{w}+\tilde{w}')\tilde{\gamma}_1.
\end{align*}
Since $\tilde{\gamma_1}\leq 0$, and by assumption $\tilde{w}+\tilde{w}'\leq \tilde{\gamma}_1+\tilde{\gamma}_2=0$, the minimum value is achieved by taking the maximum value for $\tilde{\gamma}_1$, i.e., $\tilde{\gamma}_1=-r\alpha(\epsilon_p)$ and $\tilde{\gamma}_2=r\alpha(\epsilon_p)$. Consequently, $\gamma_1=c-r\alpha(\epsilon_p)$ and $\gamma_2=c+r\alpha(\epsilon_p)$ are the optimal output values. Furthermore, from Equations \eqref{eq:th:1:JD:1} and  \eqref{eq:th:1:JD:2}, we have:
\begin{align}
    &\label{eq:th:1:opt_dist}
    P^*_{1,1}= \frac{2c-w-w'}{2r\alpha(\epsilon_p)}, \quad P^*_{1,2}= \frac{1}{2}+\frac{w'-c}{2r\alpha(\epsilon_p)}
    \\&
    \label{eq:th:1:opt_dist_2}
    P^*_{2,1}= \frac{1}{2}+\frac{w-c}{2r\alpha(\epsilon_p)},\qquad P^*_{2,2}=0.
\end{align}
To find upper-bound on the MSE of \textsc{CorBinQ}, we evaluate the total variation distance between $(P_{i,j})_{i,j\in \{1,2\}}$ and $(P^*_{i,j})_{i,j\in \{1,2\}}$, where 
\begin{align*}
    &P_{i,j}=P(Q_1(w,\mathbf{Z})=\gamma_i, Q_2(w'
,\mathbf{Z})=\gamma_j).
    \end{align*}
Note that by construction, from Algorithm \ref{Alg:algCQ}, if $\mathbf{T}_1\neq \mathbf{T}_2$,
we have: 
\begin{align*}
    \!\!P_{1,1} &= P({\mathbf{T}}_1\prec \mathbf{Z}, \mathbf{Z}\prec {\mathbf{T}}_2)
    +P(\mathbf{Z}={\mathbf{T}}_1)P(U=-1)\!+\!  P(\mathbf{Z}= {\mathbf{T}}_2)P(U'=1)\\
    &= \frac{1}{2^d}(-\floor{2^dq_1}+\floor{2^d(1-q_2)}-1+1-2^dq_1
     +\floor{2^dq_1}+2^d(1-q_2)-\floor{2^d(1-q_2)})
    \\&= 1-q_1-q_2=P^*_{1,1}.
\end{align*}
Otherwise, if $\mathbf{T}_1=\mathbf{T}_2$, then we have
$P^*_{1,1}\in [0,\frac{1}{2^d}]$, and:
\begin{align*}
   \!\!P_{1,1} &= P({\mathbf{T}}_1\prec \mathbf{Z}, \mathbf{Z}\prec {\mathbf{T}}_2)
     +P(\mathbf{Z}={\mathbf{T}}_1=\mathbf{T}_2)P(U=-1)P(U'=1)\\ 
     &= \frac{1}{2^d}(1-2^dq_1+\floor{2^dq_1})(2^d(1-q_2)-\floor{2^d(1-q_2)})
    \Rightarrow 0\leq P_{1,1}\in [0,\frac{1}{2^d}].
\end{align*}
In particular, note that:
\begin{align*}
\Rightarrow |P_{1,1}-P^*_{1,1}|\leq \frac{1}{2^d}. 
\end{align*}
Similarly, it can be observed that:
\begin{align*}
|P_{i,j}-P^*_{i,j}| \leq \frac{1}{2^d}, \quad i,j\in \{1,2\}. 
\end{align*}
So,
\begin{align*}
    &m(Q_1,Q_2,w,w')-m(Q^*_1,Q^*_2,w,w')
    \\&= \sum_{i,j\in\{1,2\}} (P_{i,j}-P^*_{i,j})(\gamma_i+\gamma_j-w-w')^2
    \leq \frac{4}{2^d} (4r\alpha(\epsilon_p))^2= \frac{r^2\alpha^2(\epsilon_p)}{2^{d-6}}.  
\end{align*}
This completes the proof for Case 1. 

\noindent\textbf{Case 2:} $|\tilde{\gamma_1}|\leq |\tilde{\gamma_2}|$ and $\tilde{w}+\tilde{w}'> \tilde{\gamma}_1+\tilde{\gamma}_2$:
\\Note that in this case, we have:
\begin{align*}
     \frac{\tilde{\gamma}_2-\tilde{w}}{\tilde{\gamma}_2-\tilde{\gamma}_1}+ \frac{\tilde{\gamma}_2-\tilde{w}'}{\tilde{\gamma}_2-\tilde{\gamma}_1}\leq 1.
\end{align*}
Furthermore, $4\tilde{\gamma}_1^2\leq 4\tilde{\gamma}_2^2$. So, Equation \eqref{eq:th:1:opt:tilde} is minimized by taking:
\begin{align*}
    &P^*_{1,1}= 0,
     \qquad \qquad  P^*_{1,2}= \frac{\tilde{\gamma}_2-\tilde{w}}{\tilde{\gamma}_2-\tilde{\gamma}_1},
   \\&
    P^*_{2,1} = \frac{\tilde{\gamma}_2-\tilde{w}'}{\tilde{\gamma}_2-\tilde{\gamma}_1},
 \qquad \qquad P^*_{2,2}=1-\frac{\tilde{\gamma}_2-\tilde{w}}{\tilde{\gamma}_2-\tilde{\gamma}_1}-\frac{\tilde{\gamma}_2-\tilde{w}'}{\tilde{\gamma}_2-\tilde{\gamma}_1} .
\end{align*}
Consequently, the optimization \eqref{eq:th:1:opt:tilde} can be rewritten as:
\begin{align}
    \nonumber&\frac{-\tilde{\gamma}_1-\tilde{\gamma}_2+\tilde{w}+\tilde{w}'}{\tilde{\gamma}_2-\tilde{\gamma}_1}(4\tilde{\gamma}^2_2)
    +\frac{2\tilde{\gamma}_2-\tilde{w}-\tilde{w}'}{\tilde{\gamma}_2-\tilde{\gamma}_1}(\tilde{\gamma}_1+\tilde{\gamma}_2)^2
    \\&\nonumber
    = -2\tilde{\gamma}_2(\tilde{\gamma}_1+\tilde{\gamma}_2)+(\tilde{w}+\tilde{w}')(3\tilde{\gamma}_2+\tilde{\gamma}_1)
\end{align}
The derivative with respect to $\tilde{\gamma}_1$ is always non-positive. So, $\tilde{\gamma}_1$ is optimized by taking $|\tilde{\gamma}_1|=|\tilde{\gamma}_2|$. Thus the optimization reduces to:
\begin{align*}
(\tilde{w}+\tilde{w}')(2\tilde{\gamma}_2),
\end{align*}
which is minimized by taking $\tilde{\gamma}_1=-r\alpha(\epsilon_p)$ and $\tilde{\gamma}_2=r\alpha(\epsilon_p)$. Consequently, $\gamma_1=c-r\alpha(\epsilon_p)$ and $\gamma_2=c+r\alpha(\epsilon_p)$. The rest of the proof for Case 2 follows by similar arguments as in Case 1 and is omitted for brevity.
\qed 
\section{Proof of Theorem \ref{prop:util}}
\label{App:prop:util}
   We have: 
\begin{align*}
 &\mathbb{E}((\overline{{W}}_{g,j}-\frac{1}{n}\sum_{i\in [n]} {w}_{i,j})^2)
 = \frac{1}{n^2}\sum_{i\in [n]} Var(\overline{W}_{i,j})+\frac{1}{n^2}\sum_{i,i'\in [n], i\neq i'} Cov(\overline{W}_{i,j},\overline{W}_{i',j}).
\end{align*}
Following \cite[Lemma 3]{sun2020ldp}, we have:
\begin{align*}
 \frac{1}{n^2}\sum_{i\in [n]} Var(\overline{W}_{i,j})\leq \frac{r^2\alpha^2(\epsilon_p)}{n}.   
\end{align*}
On the other hand:
\begin{align*}
    &\frac{1}{n^2}\sum_{i,i'\in [n], i\neq i'} Cov(\overline{W}_{i,j},\overline{W}_{i',j})=\frac{1}{2n^2} \sum_{i\in [n]} Cov(\overline{W}_{i,j},\overline{W}_{p_i,j}),
\end{align*}
where $p_i$ is the index of the client paired with $\mathcal{C}_i$ as described in Algorithm \ref{Alg:CorBinFL}. Furthermore,
\begin{align*} &Cov(\overline{W}_{i,j},\overline{W}_{p_i,j})= P_{1,1}(-r\alpha(\epsilon_p))^2
+ (P_{1,2}+P_{2,1})(-r\alpha(\epsilon_p))(r\alpha(\epsilon_p))
+P_{2,2}(r\alpha(\epsilon_p))^2
\\&- ({w}_{i,j}-c)({w}_{p_i,j}-c).
\end{align*}
Let us assume that ${w}_{i,j}+{w}_{p_i,j}\leq 2c$. Then, from Equations \eqref{eq:th:1:opt_dist} and \eqref{eq:th:1:opt_dist_2}, as $d\to \infty$, we have:
\begin{align*} &Cov(\overline{W}_{i,j},\overline{W}_{p_i,j})= \left(\frac{2c-w_{i,j}-w_{p_i,j}}{2r\alpha(\epsilon_p)}\right)(-r\alpha(\epsilon_p))^2
\\&+ \left(1-\frac{2c-w_{i,j}-w_{p_i,j}}{2r\alpha(\epsilon_p)}\right)(-r\alpha(\epsilon_p))(r\alpha(\epsilon_p))
- ({w}_{i,j}-c)({w}_{p_i,j}-c)
\\& = r^2\alpha^2(\epsilon_p)\left(\frac{2c-w_{i,j}-w_{p_i,j}}{r\alpha(\epsilon_p)}-1\right)-
({w}_{i,j}-c)({w}_{p_i,j}-c)
\\& \leq r^2\alpha^2(\epsilon_p)\left(\frac{2}{\alpha(\epsilon_p)}-1\right)- r^2
 = r^2(2\alpha(\epsilon_p)-\alpha^2(\epsilon_p)-1)
 = -r^2(\alpha(\epsilon_p)-1)^2.
\end{align*}
Consequently,
\begin{align*}
  &  \lim_{d\to \infty}\mathbb{E}((\overline{{w}}_{g,j}-\frac{1}{n}\sum_{i\in [n]} {w}_{i,j})^2)
    \leq \frac{1}{n}r^2\alpha^2(\epsilon_p)- \frac{1}{2n} r^2(\alpha(\epsilon_p)-1)^2
    =\frac{1}{2n}r^2(2\alpha^2(\epsilon_p)-(\alpha(\epsilon_p)-1)^2) 
    \\&= \frac{1}{2n}r^2((\sqrt{2}+1)\alpha(\epsilon_p)-1)((\sqrt{2}-1)\alpha(\epsilon_p)+1).
\end{align*}
This completes the proof for the case where $\overline{w}_{i,j}+\overline{w}_{p_i,j}\leq 2c$. The proof for the case where  $\overline{w}_{i,j}+\overline{w}_{p_i,j}> 2c$ follows by a similar argument and is omitted.
\qed

\section{Proof of Theorem \ref{th:3}}
We provide an outline of the proof. Let the local update at client $\mathcal{C}_i$ be denoted by $\mathbf{w}_i\in [\mathbf{c}-\mathbf{r},\mathbf{c}+\mathbf{r}]$, and $\overline{\mathbf{W}}_i$ be the obfuscated local update after applying the obfuscation step in AugCorBin-FL. Furthermore, let $\mathcal{T}\subseteq \mathbb{R}^m$. We wish to find $(\epsilon_u,\delta)$ such that
\begin{align*}
    P(\frac{1}{n} \sum_{i\in [n]} \mathbf{\overline{W}}_i\in \mathcal{T})\leq e^{\epsilon_u}P(\frac{1}{n} \sum_{i\in [n], i\neq 1} \mathbf{\overline{W}}_i\in \mathcal{T})+\delta.
\end{align*}
We first prove the result for the case when $\mathcal{C}_1$ is among the $\gamma$ fraction of clients using the LDP-FL mechanism. Let $\mathcal{A}\subseteq [n]$ be the set of indices of clients $\mathcal{C}_i$ which belong to the $\gamma$ fraction of clients using the LPD-FL mechanism. Then, 
\begin{align*}
\sum_{i\in [n]-\{1\}} \mathbf{\overline{W}}_i
= \sum_{i\in \mathcal{A}-\{1\}} \mathbf{\overline{W}}_i+\sum_{i\in \mathcal{A}^c} \mathbf{\overline{W}}_i
\end{align*}
Let us define $\mathbf{B}= \sum_{i\in \mathcal{A}-\{1\}} \mathbf{\overline{W}}_i$ and $\mathbf{B}'=\sum_{i\in \mathcal{A}^c} \mathbf{\overline{W}}_i$. Then, $\mathbf{B}$ and $\mathbf{B}'$ are independent of each other. 
Note that for $i\in \mathcal{A}, j\in [m]$,  the LDPQ quantizer generates an output $\overline{W}_{i,j}$ which is a binary with $P(\overline{W}_{i,j}=c+r\alpha(\epsilon_p))= \frac{1}{2}+\frac{w-c}{2r\alpha(\epsilon_p)}$. 
Let us define $V_{j}=\frac{1}{2}+\frac{\overline{W}_{i,j}-c}{2r\alpha(\epsilon_p)}$.
Then, $P(V_{j}=1)=P(\overline{W}_{i,j}=c+r\alpha(\epsilon_p))\in [\frac{1}{2}- \frac{1}{2\alpha(\epsilon_p)}, \frac{1}{2}+\frac{1}{2\alpha(\epsilon_p)}]$. Consequently, $\frac{1}{2}+\frac{B_{j}-c}{2r\alpha(\epsilon_p)}$ is Binomial with parameters $(\gamma n-1,p)$, where $p\in [\frac{1}{2}- \frac{1}{2\alpha(\epsilon_p)}, \frac{1}{2}+\frac{1}{2\alpha(\epsilon_p)}]$. The differential privacy guarantees follow by \cite[Theorem 1]{agarwal2018cpsgd}. To elaborate, following the notation of \cite[Theorem 1]{agarwal2018cpsgd}, we define:
\begin{align*}
    \\&f(\mathcal{D}_1)= \frac{1}{\mathbf{r}\alpha(\epsilon_p)} (\overline{\mathbf{W}}_{1}-\mathbf{c})\in \{-1,1\},
    \\&\mathbf{Z}= \frac{1}{2}+\frac{\mathbf{B}-\mathbf{c}}{2\mathbf{r}\alpha(\epsilon_p)}\sim \textsc{Binomial}(N, p),
    \\& N= \gamma n-1,
    \qquad  p=\frac{1}{2}-\frac{1}{2\alpha(\epsilon_p)},\quad s=2,
    \\& \mathcal{M}_b^{N,p,s}=
    f(\mathcal{D}_1)+(\mathbf{Z}-Np)s
    \\&= \frac{1}{\mathbf{r}\alpha(\epsilon_p)}(\mathbf{\overline{W}}_1+\mathbf{B})-n\gamma(1-\frac{1}{\alpha(\epsilon_p)})+1-\frac{3\mathbf{c}}{2\mathbf{r}\alpha(\epsilon_p)},
    \\& \Delta_1=\max_{\mathbf{w}_{1},\mathbf{w}'_1} \|\mathbf{\overline{W}}_1- \mathbf{\overline{W}}'_1\|_1\leq   2mr\alpha(\epsilon_p)
     \\&
\Delta_2=\max_{\mathbf{w}_{1},\mathbf{w}'_1} \|\mathbf{\overline{W}}_1- \mathbf{\overline{W}}'_1\|_2\leq  2\sqrt{m}r\alpha(\epsilon_p)
\\&  \Delta_\infty=\max_{\mathbf{w}_{1},\mathbf{w}'_1} \|\mathbf{\overline{W}}_1- \mathbf{\overline{W}}'_1\|_\infty\leq 2r\alpha(\epsilon_p).
\end{align*}
The proof then follows by substituting the above variables into Equation (7) in \cite{agarwal2018cpsgd}.

Similarly, if $\mathcal{C}_1$ belongs to the $(1-\gamma)$ fraction of clients using the CorBin-FL mechanism.
Then, we first note that:
\begin{align*}
 &P(\frac{1}{n} \sum_{i\in [n]} \mathbf{\overline{W}}_i\in \mathcal{T})
=\!\!\!\! \sum_{\mathbf{\overline{w}}_{p_1}\in \mathcal{W}} 
\!\!
\!\!
P_{\mathbf{\overline{W}}_{p_1}}(\mathbf{\overline{w}}_{p_1})
P(\frac{1}{n}
\!\!\sum_{\substack{i\in [n]-\{p_1\}}}
\!\!\!\!\!\!\!\!
\mathbf{\overline{W}}_i\in \mathcal{T}'_{\mathbf{w}_{p_1}}| \mathbf{\overline{W}}_{p_1}= \mathbf{\overline{w}}_{p_1})
\\&= \!\!\!\!\sum_{\mathbf{\overline{w}}_{p_1}\in \mathcal{W}}\!\!\!\! 
P_{\mathbf{\overline{W}}_{p_1}}(\mathbf{\overline{w}}_{p_1})
P(\frac{1}{n}(\mathbf{\overline{W}'}_1+
\!\!\!\!\sum_{i\in [n]-\{1,p_1\}} 
\!\!\!\!
\mathbf{\overline{W}}_i)\in \mathcal{T}'_{\mathbf{w}_{p_1}}),
\end{align*}
where $\mathcal{W}= \{(c_j+u_jr_j\alpha(\epsilon_p))_{j\in [m]}| u_j\in \{-1,1\}, j\in [m] \}$,
$\mathcal{T}'_{\mathbf{w}_{p_1}}= \mathcal{T}-\frac{1}{n}\mathbf{w}_{p_1}$ is a Borel set, $\mathbf{\overline{W}}'_{1}$ is a random vector with distribution $P_{\mathbf{\overline{W}}_1|\mathbf{\overline{W}}_{p_1}}(\cdot|\mathbf{\overline{w}}_{p_1})$, and in the last equality, we have used the fact that in AugCorBin-FL obfuscated updates are only pairwise dependent.  Similarly, 
\begin{align*}
&P(\frac{1}{n} \sum_{i\in [n], i\neq 1} \mathbf{\overline{W}}_i\in \mathcal{T})
=\sum_{\mathbf{\overline{w}}_{p_1}\in \mathcal{W}}P_{\mathbf{\overline{W}}_{p_1}}(\mathbf{\overline{w}}_{p_1})P(\frac{1}{n} \sum_{i\in [n]- \{1,p_1\}} \mathbf{\overline{W}}_i\in \mathcal{T}'_{\mathbf{w}_{p_1}})
\end{align*}
Consequently, it suffices to find $(\epsilon_u,\delta)$ such that for all 
$\mathbf{\overline{w}}_{p_1}\in \mathcal{W}$, we have:
\begin{align*}
 &   P(\frac{1}{n}(\mathbf{\overline{W}}'_i+\sum_{i\in [n]-\{1,p_1\}} \mathbf{\overline{W}}_i)\in \mathcal{T}_{\mathbf{\overline{w}}_{p_1}})
    \leq e^{\epsilon_u}P(\frac{1}{n} \sum_{i\in [n]-\{1,p_1\}} \mathbf{\overline{W}}_i\in \mathcal{T}_{\mathbf{\overline{w}}_{p_1}})+\delta.
\end{align*}

\begin{figure*}[t]
    \centering
    \begin{subfigure}[b]{0.32\textwidth}
        \centering
        \includegraphics[width=\textwidth]{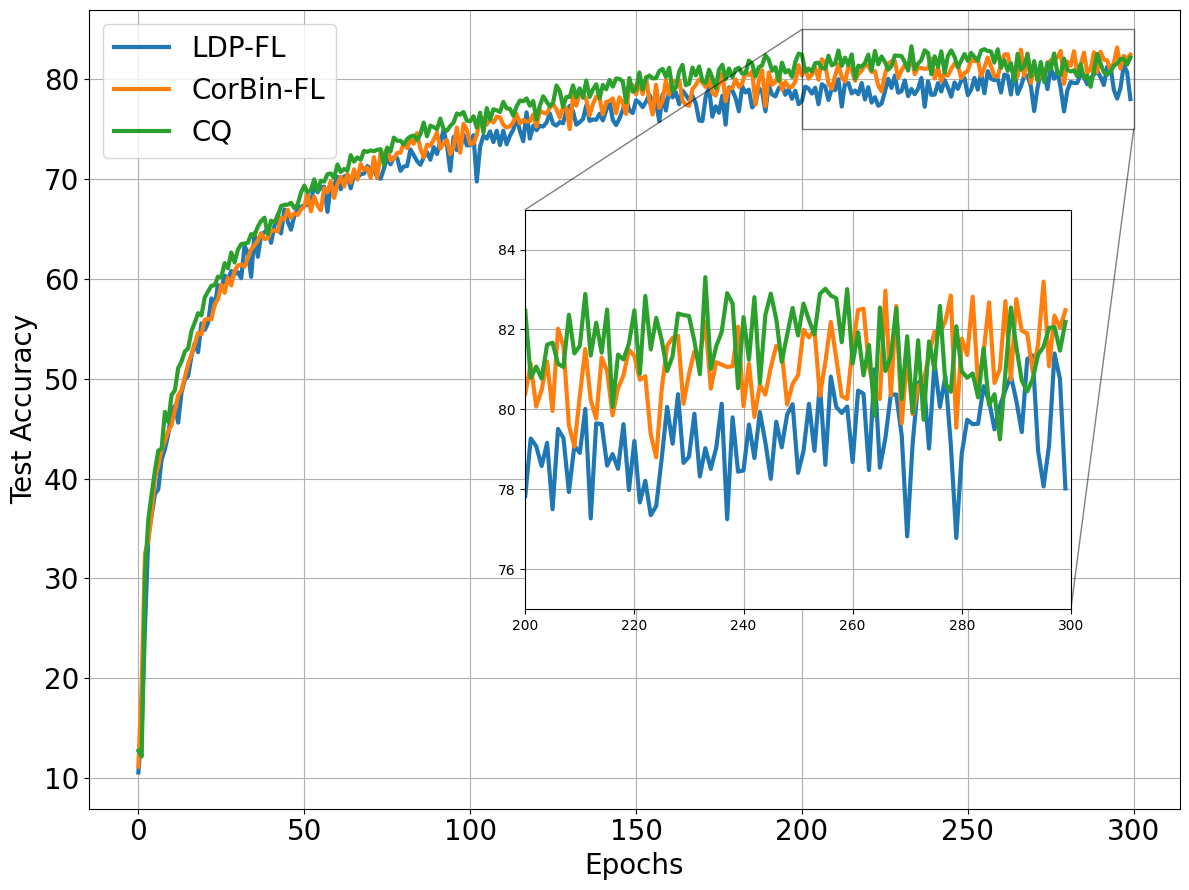}
        \caption{}
        \label{fig: compare_with_Suresh}
    \end{subfigure}
    \hfill
    \begin{subfigure}[b]{0.32\textwidth}
        \centering
        \includegraphics[width=\textwidth]{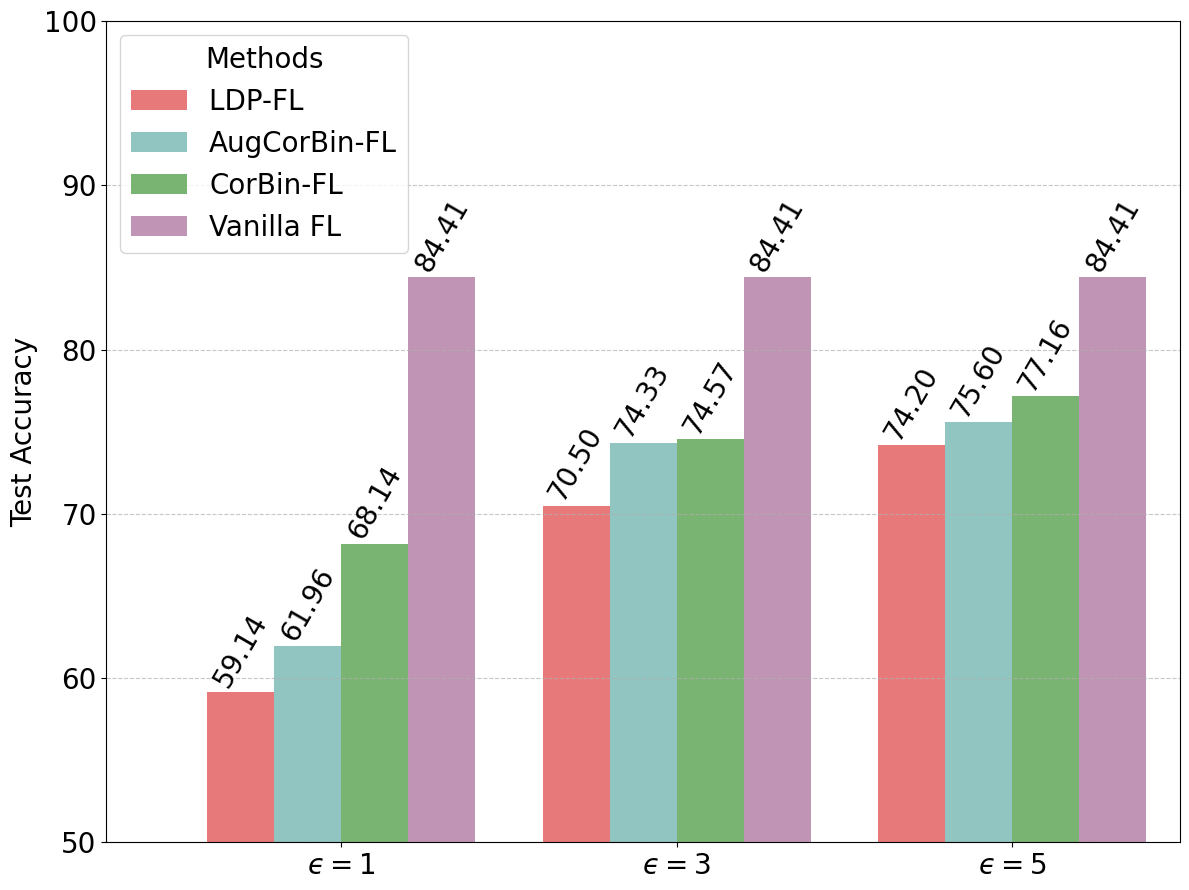}
        \caption{}
        \label{fig: different_ep}
    \end{subfigure}
    \hfill
    \begin{subfigure}[b]{0.32\textwidth}
        \centering
        \includegraphics[width=\textwidth]{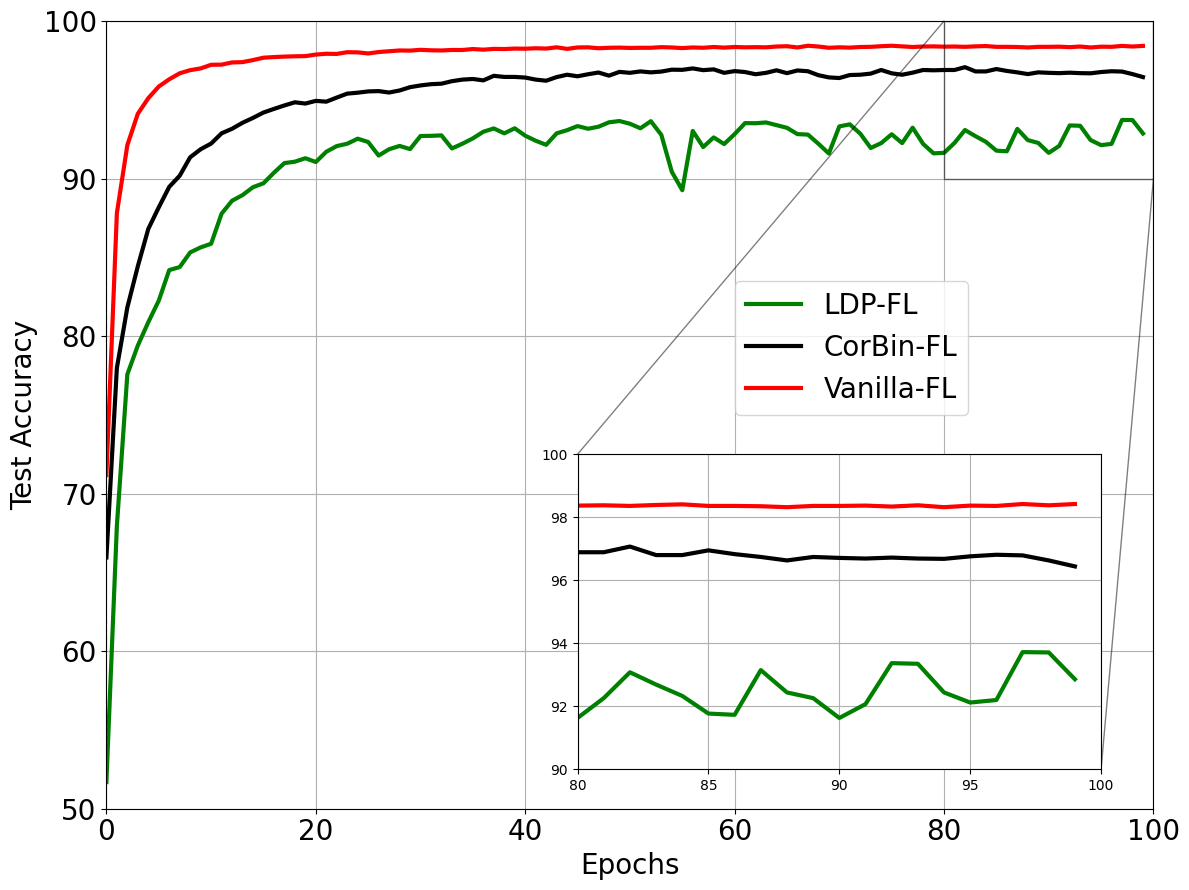}
        \caption{}
        \label{fig: mnist}
    \end{subfigure}
    \caption{Additional experimental results evaluating (a) the accuracy performance of LDP-FL, CorBin-FL, and CQ \cite{suresh2022correlated}, (b) Performance of LDP-FL, AugCorBin-FL, and CorBin-FL under various PLDP privacy budgets, and (c) accuracy performance of LDP-FL and CorBin-FL on the MNIST dataset.}
    \label{fig:all_comparisons_2}
\end{figure*}

The privacy guarantees then follow by applying \cite[Theorem 1]{agarwal2018cpsgd} as in the previous case.  The mean square error guarantees follow directly from Proposition \ref{prop:1} and Theorem \ref{th:1}.
\section{Proof of Theorem \ref{prop:4}}
    We provide an outline of the proof. Let $\mathcal{P}_i, i\in [\frac{n}{2}]$ be the collection of pairings of the clients in CorBin-FL, and let $E_i$ be the indicator that exactly one of the two clients in the $i$th pair drops out. Then, 
    \begin{align*}
      &  P(E_i=1)= 2p(1-p),
      \quad \mathbb{E}(\sum_{i\in [\frac{n}{2}]}E_i) = np(1-p)
      \\& Var(\sum_{i\in [\frac{n}{2}]}E_i)= np(1-p)(1-2p(1-p)).
    \end{align*}
    Consequently, using the Chebychev's inequality, we have:
    \begin{align*}
        P(\sum_{i\in [\frac{n}{2}]}E_i\in [\gamma n,\gamma' n])\geq 1-(\delta-\delta_1)
    \end{align*}
    Let $\mathcal{E}$ represent the event that $\sum_{i\in [\frac{n}{2}]}E_i\in [\gamma n,\gamma' n]$. Then, 
    \begin{align*}
        &P(\sum_{i\in [n]}\overline{W}_i\leq \mathcal{T}) \leq P(\mathcal{E})P(\sum_{i\in [n]}\overline{W}_i\leq \mathcal{T}|\mathcal{E})+P(\mathcal{E}^c)
      \leq P(\sum_{i\in [n]}\overline{W}_i\leq \mathcal{T}|\mathcal{E})+(\delta-\delta_1)
        \\&\leq e^{\epsilon_u}P(\sum_{i=2}^n\overline{W}_i\leq \mathcal{T}|\mathcal{E})+\delta_1+\delta-\delta_1,
    \end{align*}
    where the last inequality follows from the proof of Theorem \ref{th:3}. This completes the proof of the UCDP guarantee. Furthermore, let $N= \sum_{i\in [n]}F_i$, where $F_i$ is the indicator that the $i$th client does not dropout. 
    Note that:
        \begin{align*}
      &  P(F_i=1)= p,
     \qquad \mathbb{E}(\sum_{i\in [n]}F_i) = np,
   \qquad Var(\sum_{i\in [n]}F_i)= np(1-p).
    \end{align*}
    Then, 
    \begin{align*}
        P(\sum_{i\in [n]}F_i\in [\theta n,\theta' n])\geq 1-(\delta-\delta_1),
    \end{align*}
    where $\theta=p(1-p)-\sqrt{\frac{p(1-p)}{(\delta-\delta_1)n}}$ and $\theta'=p(1-p)+\sqrt{\frac{p(1-p)}{(\delta-\delta_1)n}}$. Let $\mathcal{F}$ be the event that $\sum_{i\in [n]}F_i\in [\theta n,\theta' n]$, then:
    
    \begin{align*}
     &\lim_{d\to \infty}\mathbb{E}((\overline{{W}}_{g,j}-\frac{1}{N}\sum_{i:F_i=1} {w}_{i,j})^2)
     \\& \leq 
     \lim_{d\to \infty}\mathbb{E}((\overline{{W}}_{g,j}-\frac{1}{N}\sum_{i:F_i=1} {w}_{i,j})^2|\mathcal{E}\cap \mathcal{F})
      + (2r\alpha(\epsilon_p))^2(P(\mathcal{E}^c)+P(\mathcal{F}^c))
     \\&\leq
       \frac{(1-\gamma)}{2n\theta}r^2\left((\sqrt{2}-1)\alpha(\epsilon_p)+1\right)\left(  (\sqrt{2}+1)\alpha(\epsilon_p)-1\right)
        +\frac{\gamma' r^2\alpha^2(\epsilon_p)}{n\theta}+ 8r^2\alpha^2(\epsilon_p)\delta_1.
    \end{align*}
    This completes the proof. 
    \qed
    
\section{Additional Experimental Results}
In this section, we provide additional empirical evaluations and experimental results to evaluate various aspects of the proposed privacy mechanisms in comparison with existing mechanisms. 
\subsubsection{Experiment 5 - Correlated quantization without privacy constraints:}
An advantage of CorBin-FL is the low communication cost due to one-bit quantization of each model parameter. Other works have considered correlated quantization without privacy constraints and towards reducing the communication overhead. In this experiment, we take the correlated quantization (CQ) method of \cite{suresh2022correlated} as a representative example, and compare its accuracy performance with that of CorBin-FL and LDP-FL when $\epsilon_p=10$, i.e. very weak privacy constraints. 
We perform the simulation over the CIFAR-10 dataset, as illustrated in Figure \ref{fig: compare_with_Suresh}. The experiment involves 50 clients in a federated learning setting, with a batch size of 64. For CorBin-FL, we set the PLDP budget to $\epsilon_p = 10$ and use $d = 5$ common random bits. The global learning rate hyperparameter $\lambda$ equal to one for all methods. We include one-bit correlated quantization method \cite{suresh2022correlated} without privacy guarantees, as well as LDP-FL with a significantly relaxed privacy budget of $\epsilon_p = 50$ for comparison. The results demonstrate that CorBin-FL with $\epsilon_p=10$ achieves accuracy that is almost equal to that of the CQ with $\epsilon_p\to \infty$ and LDP-FL with $\epsilon_p=50$.

\subsubsection{Experiment 6 - Accuracy  for Varying Privacy Budgets:}
We evaluate the accuracy performance of CorBin-FL, AugCorBin-FL, and LDP-FL on the CIFAR10 dataset, considering various privacy budgets $\epsilon_p \in \{1, 3, 5\}$. We perform a grid search over  $\lambda \in \{0.1, 0.2, ..., 1.0\}$ for each value of $\epsilon_p$. Figure \ref{fig: different_ep} presents the best accuracy achieved by each method for the different privacy budgets.

\subsubsection{Experiment 7- Experiments on the MNIST Dataset:}
We conduct experiments on the MNIST dataset with a fixed privacy budget of $\epsilon_p = 0.5$ and using a two-layer VGG model. We compare the performance of CorBin-FL, LDP-FL, and the Vanilla-FL with no privacy guarantees. 
We perform a grid search over $\lambda \in \{0.1, 0.2, ..., 1.0\}$ for all methods to ensure fair comparison, the result is shown in Figure \ref{fig: mnist}. The results demonstrate that CorBin-FL achieves better accuracy compared to LDP-FL and the gap becomes larger for smaller values of epsilon. The accuracy loss of CorBin-FL compared to vanilla-FL is less than 1.5\% at $\epsilon_p = 0.5$.

\begin{figure*}[t]
\centering
\includegraphics[width=0.9\textwidth]{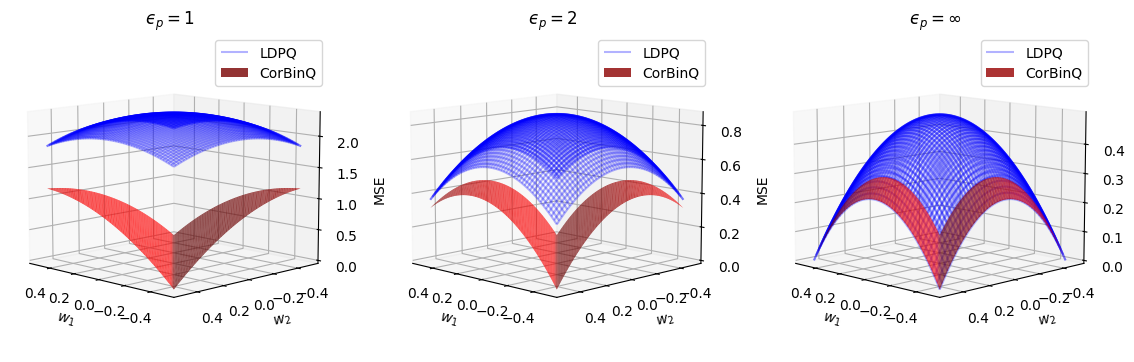} 
\caption{MSE of CorBinQ and LDPQ pairs of quantizers for inputs $w_1,w_2\in[-0.5,0.5]$.}
\label{fig: fig1}
\end{figure*}

\subsubsection{Experiment 8 - MSE comparison of LDPQ and CorBinQ:}
We focus on the quantizers used in the CorBin-FL and LDP-FL mechanisms, namely the CorBinQ and LDPQ quantizers. We plot the resulting MSE when feeding a pair of quantziers with two input weights $w_1,w_2\in [c-r,c+r]$, where we have taken $c=0$ and $r=0.5$. Figure~\ref{fig: fig1} show the MSE comparison between a pair of clients utilizing CorBinQ in Algorithm \ref{Alg:algCQ} and a pair of clients using the LDPQ in Algorithm \ref{Alg:algQ}. The figure demonstrates the MSE gains of CorBinQ, particularly at smaller values of the privacy budget $\epsilon_p$. There is a larger gap in variance between CorBinQ and LDPQ methods as $\epsilon_p$ decreases. This suggests that correlated randomness is especially beneficial in scenarios requiring stronger privacy guarantees (i.e., smaller $\epsilon_p$ values).

\end{appendices}

\end{document}